\documentclass[10pt,twocolumn,letterpaper]{article}

\usepackage{cvpr}              

\usepackage{graphicx}
\usepackage{amsmath}
\usepackage{amssymb}
\usepackage{booktabs}
\usepackage{multirow}
\usepackage{makecell}
\usepackage{floatrow}
\usepackage{caption}
\usepackage{xcolor}
\definecolor{citecolor}{HTML}{2980b9}
\definecolor{linkcolor}{HTML}{c0392b}
\usepackage{stfloats}

\usepackage{adjustbox}
\usepackage{colortbl}

\makeatletter
  \newcommand\figcaption{\def\@captype{figure}\caption}
  \newcommand\tabcaption{\def\@captype{table}\caption}
\makeatother
\usepackage[hidelinks,pagebackref=true,breaklinks=true,colorlinks,bookmarks=false,citecolor=citecolor,letterpaper=true,linkcolor=linkcolor]{hyperref}

\usepackage[capitalize]{cleveref}
\crefname{section}{Sec.}{Secs.}
\Crefname{section}{Section}{Sections}
\Crefname{table}{Table}{Tables}
\crefname{table}{Tab.}{Tabs.}
\newcommand\blfootnote[1]{%
  \begingroup
  \renewcommand\thefootnote{}\footnote{#1}%
  \addtocounter{footnote}{-1}%
  \endgroup
}

\def\cvprPaperID{1503} 
\def\confName{CVPR}
\def\confYear{2023}

\begin{document}

\title{Parameter is Not All You Need:\\Starting from Non-Parametric Networks for 3D Point Cloud Analysis}

\author{Renrui Zhang$^{1,5}$, Liuhui Wang$^{2,6}$, Ziyu Guo$^{2}$, Yali Wang$^{4,5}$, Peng Gao$^{5}$, \vspace{0.3cm} Hongsheng Li$^{1}$, Jianbo Shi$^{\dagger3}$ \\
  $^1$CUHK MMLab\quad 
  $^2$Peking University\quad
  $^3$University of Pennsylvania\\
  $^4$Shenzhen Institute of Advanced Technology, Chinese Academy of Sciences\\
  $^5$Shanghai Artificial Intelligence Laboratory\quad
  $^6$Heisenberg Robotics\vspace{0.2cm}\\
\texttt{\{zhangrenrui, gaopeng\}@pjlab.org.cn},\quad
\texttt{jshi@seas.upenn.edu}\\
\texttt{wangliuhui0401@pku.edu.cn},\quad
\texttt{hsli@ee.cuhk.edu.hk}
}

\maketitle
\blfootnote{$\dagger$ Corresponding author}
\begin{abstract}

We present a \textbf{N}on-parametric \textbf{N}etwork for 3D point cloud analysis, \textbf{Point-NN}, which consists of purely non-learnable components: farthest point sampling (FPS), $k$-nearest neighbors ($k$-NN), and pooling operations, with trigonometric functions.
Surprisingly, it performs well on various 3D tasks, requiring no parameters or training, and even surpasses existing fully trained models. Starting from this basic non-parametric model, we propose two extensions.  
First, Point-NN can serve as a base architectural framework to construct \textbf{P}arametric \textbf{N}etworks by simply inserting linear layers on top. Given the superior non-parametric foundation, the derived \textbf{Point-PN} exhibits a high performance-efficiency trade-off with only a few learnable parameters. 
Second, Point-NN can be regarded as a plug-and-play module for the already trained 3D models during inference. Point-NN captures the complementary geometric knowledge and enhances existing methods for different 3D benchmarks without re-training. 
We hope our work may cast a light on the community for understanding 3D point clouds with non-parametric methods.
Code is available at \url{https://github.com/ZrrSkywalker/Point-NN}.

\end{abstract}

\section{Introduction}
\label{sec:intro}

Point cloud 3D data processing is a foundational operation in autonomous driving~\cite{chen2017multi,navarro2010pedestrian,kidono2011pedestrian}, scene understanding~\cite{aldoma2012tutorial,verdoja2017fast,chen2019deep,zheng2013beyond}, and robotics~\cite{rusu2009close, correll2016analysis, mousavian20196}.  Point clouds contain unordered points discretely depicting object surfaces in 3D space. Unlike grid-based 2D images, they are distribution-irregular and permutation-invariant, which leads to non-trivial challenges for algorithm designs.

\begin{figure}[t]
  \centering
\includegraphics[width=0.96\textwidth]{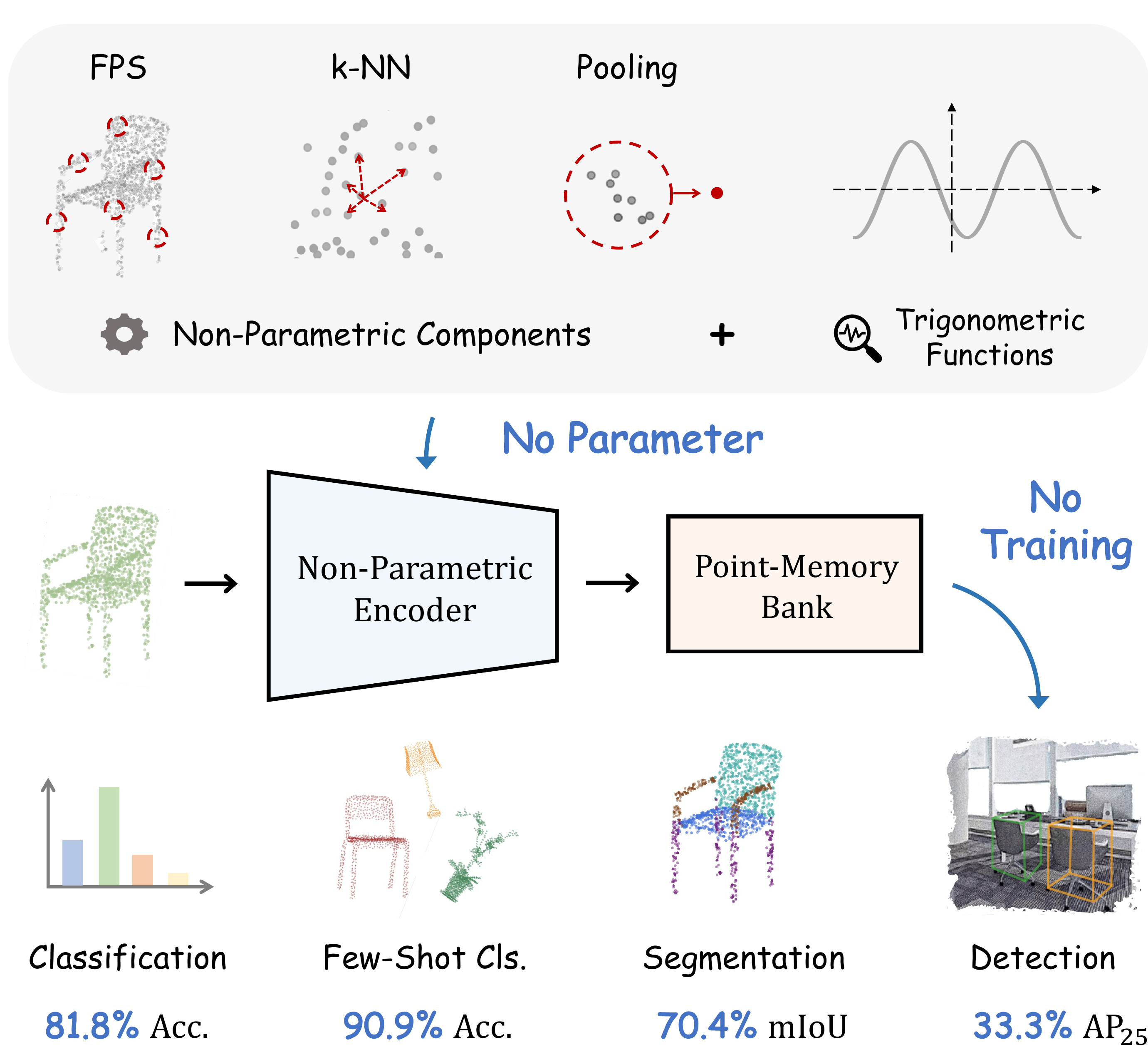}
\vspace{0.2cm}
   \caption{\textbf{The Pipeline of Non-Parametric Networks.} Point-NN is constructed by the basic non-parametric components without any learnable operators. Free from training, Point-NN can achieve favorable performance on various 3D tasks.}
    \label{fig1}
\vspace{-0.2cm}
\end{figure}

Since PointNet++~\cite{qi2017pointnet++}, the prevailing trend has been adding advanced local operators and scaled-up learnable parameters.
Instead of max pooling for feature aggregation, mechanisms are proposed to extract local geometries, e.g., adaptive point convolutions~\cite{xu2021paconv,pvcnn,thomas2019kpconv,densepoint,curvenet} and graph-like message passing~\cite{dgcnn,guo2021pct,zhao2021point}. The performance gain also rises from scaling up the number of parameters, e.g., KPConv~\cite{thomas2019kpconv}'s 14.3M and PointMLP~\cite{pointmlp}'s 12.6M, is much larger than PointNet++'s 1.7M. 
This trend has increased network complexity and computational resources. 

Instead, the non-parametric framework underlying all the learnable modules remains nearly the same since PointNet++, including farthest point sampling (FPS), $k$-Nearest Neighbors ($k$-NN), and pooling operations. Given that few works have investigated their efficacy, we ask the question: \textit{can we achieve high 3D point cloud analysis performance using only these non-parametric components?}

We present a \textbf{N}on-parametric \textbf{N}etwork, termed Point-NN, which is constructed by the aforementioned non-learnable components. Point-NN, as shown in Figure~\ref{fig1}, consists of a non-parametric encoder for 3D feature extraction and a point-memory bank for task-specific recognition. The multi-stage encoder applies FPS, $k$-NN, trigonometric functions, and pooling operations to progressively aggregate local geometries, producing a high-dimensional global vector for the point cloud. We only adopt simple trigonometric functions to reveal local spatial patterns at every pooling stage without learnable operators. Then, we adopt the non-parametric encoder of Point-NN to extract the training-set features and cache them as the point-memory bank. For a test point cloud, the bank outputs task-specific predictions via naive feature similarity matching, which validates the encoder's discrimination ability.

Free from any parameters or training, Point-NN unexpectedly achieves favorable performance on various 3D tasks, e.g., shape classification, part segmentation and 3D object detection, indicating the strength of the long-ignored non-parametric operations. Compared to existing parametric methods, Point-NN even surpasses the fully trained 3DmFV~\cite{3dmfv} by \textbf{+2.9\%} classification accuracy on OBJ-BG split of ScanObjectNN~\cite{scanobjectnn}. Especially for few-shot classification, Point-NN significantly exceeds PointCNN~\cite{li2018pointcnn} and other models~\cite{qi2017pointnet,qi2017pointnet++} by more than +20\% accuracy, indicating its superiority in low-data regimes. \textit{Starting from this simple-but-effective Point-NN, we propose two approaches by leveraging the non-parametric components to benefit 3D point cloud analysis.}

\begin{figure}[t]
  \centering
\vspace{0.1cm}
\includegraphics[width=\textwidth]{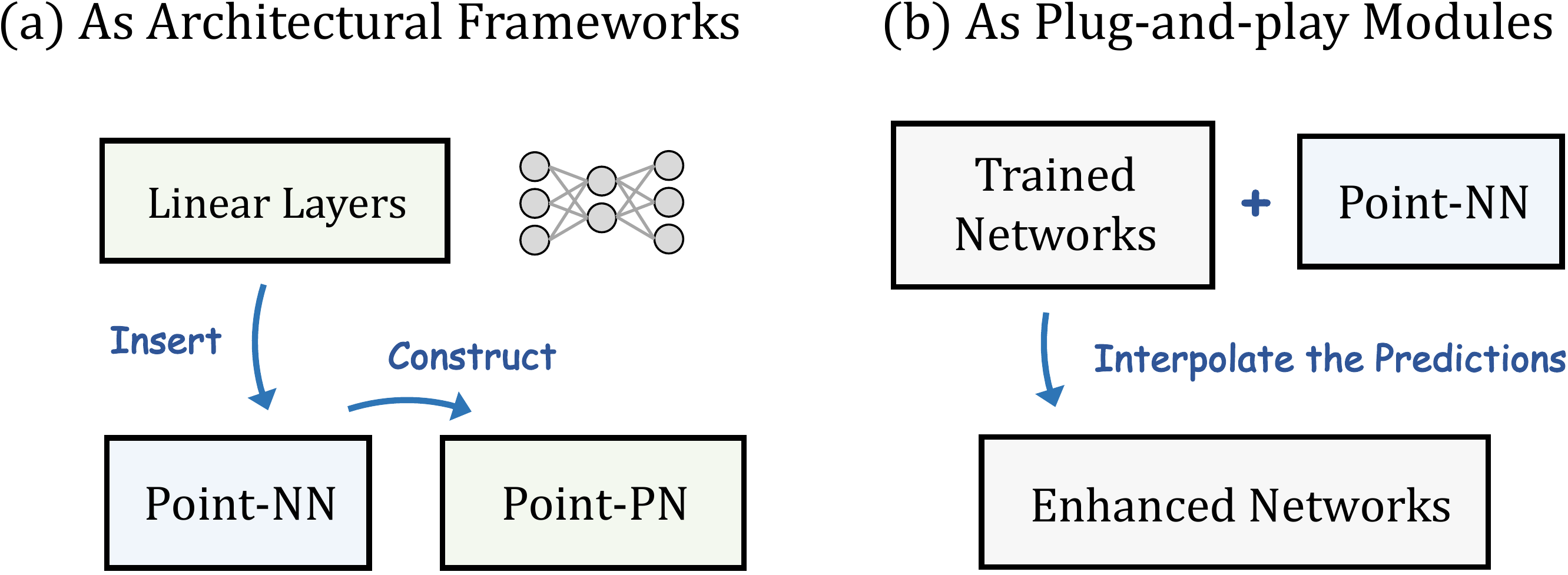}
\vspace{0.05cm}
   \caption{\textbf{Two Applications of Point-NN.} (a) As a non-parametric framework to construct parametric networks by simply inserting linear layers. (b) As a plug-and-play module to enhance already trained networks without re-training.}
    \label{fig2}
\end{figure}

First, Point-NN can serve as an architectural precursor to construct \textbf{P}arametric \textbf{N}etworks, termed as Point-PN, shown in Figure~\ref{fig2} (a). As we have fully optimized the non-parametric framework, the Point-PN can be simply derived by inserting linear layers into every stage of Point-NN. Point-PN contains no complicated local operators, but only linear layers and trigonometric functions inherited from Point-NN. Experiments show that Point-PN can achieve competitive performance with a small number of parameters, e.g., 87.1\% classification accuracy with 0.8M on the hardest split of ScanObjectNN~\cite{scanobjectnn}.

Second, Point-NN can be regarded as a plug-and-play module to enhance the already trained 3D models without re-training, as shown in Figure~\ref{fig2} (b). We directly fuse the predictions between Point-NN and off-the-shelf 3D models during inference by linear interpolation. Given the training-free manner, Point-NN mainly focuses on low-level 3D structural signals by trigonometric functions, which provides complementary knowledge to the high-level semantics of existing 3D models. On different tasks, Point-NN exhibits consistent performance boost, e.g., +2.0\% classification accuracy on ScanObjectNN~\cite{scanobjectnn} and +11.02\% detection AR$_{25}$ on ScanNetV2~\cite{ScanNetV2}.

Our contributions are summarized below:
\begin{itemize}
   \item We propose to revisit the non-learnable components in 3D networks, and, \textit{for the first time}, develop a non-parametric method, Point-NN, for 3D analysis.
   \item By using Point-NN as a basic framework, we introduce its parameter-efficient derivative, Point-PN, which exerts superior performance without advanced operators.
   \item As a plug-and-play module, the complementary Point-NN can boost off-the-shelf trained 3D models for various 3D tasks directly during inference.
\end{itemize}

\vspace{0.2cm}
\section{Non-Parametric Networks}

In this section, we first investigate the basic non-parametric components in existing 3D models (Sec.~\ref{s3.1}). Then, we integrate them into Point-NN, which consists of a non-parametric encoder (Sec.~\ref{s3.2}) and a point-memory bank (Sec.~\ref{s3.3}). Finally, we introduce how to apply Point-NN for various 3D tasks (Sec.~\ref{s3.4}).

\subsection{Basic Components}
\label{s3.1}
Underlying most of the point cloud processing networks~\cite{pointmlp,qi2017pointnet++} are a series of non-parametric components, i.e., farthest point sampling (FPS), $k$-Nearest Neighbors ($k$-NN), and pooling operations. These building blocks are non-learnable during training and iteratively stacked into multiple stages to construct a pyramid hierarchy. 

\begin{figure}[t]
  \centering
\includegraphics[width=\textwidth]{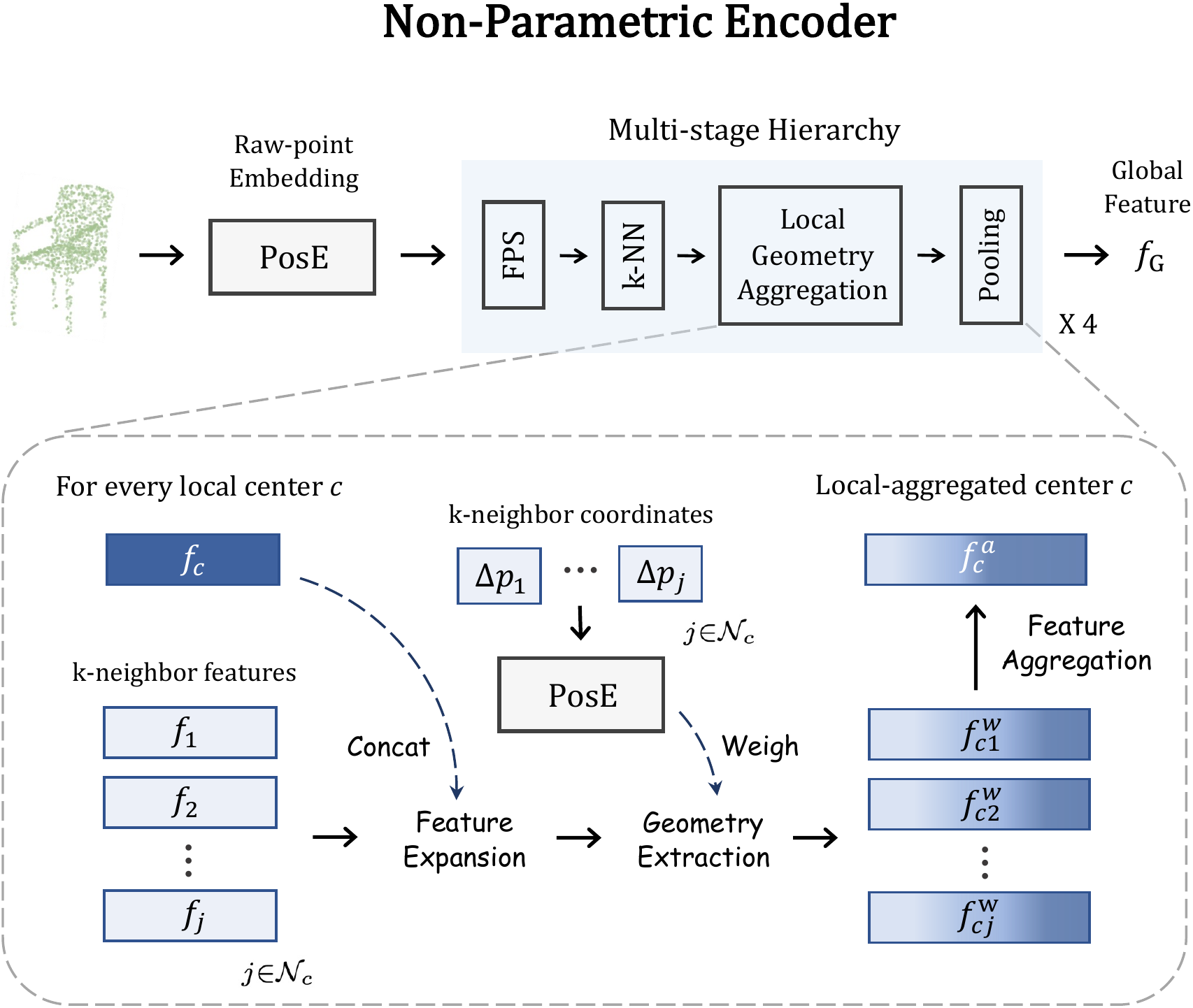}
   \caption{\textbf{Non-Parametric Encoder of Point-NN.} We first utilize trigonometric functions, denoted as $\operatorname{PosE}(\cdot)$, to encode raw points into high-dimensional vectors, and then adopt non-learnable operations to hierarchically aggregate local features.}
    \label{fig3}
\end{figure}

For each stage, we denote the input point cloud representation from the last stage as $\{p_i, f_i\}_{i=1}^M$, where $p_i \in \mathbb{R}^{1 \times 3}$ and $f_i \in \mathbb{R}^{1 \times C}$ denote the coordinate and feature of point $i$. First, FPS is adopted to downsample the point number from $M$ to $\frac{M}{2}$ by selecting a subset of local center points,
\begin{align}
    \{p_c, f_c\}_{c=1}^{\frac{M}{2}} = \operatorname{FPS}\Big(\{p_i, f_i\}_{i=1}^M\Big).
\end{align}
Then, $k$-NN, or ball query, is responsible for grouping $k$ spatial neighbors for each center $c$ from the original $M$ points, which forms the local 3D region,
\begin{align}
    {\cal{N}}_c = \operatorname{\textit{k}-NN}\Big(p_c,\ \{p_i\}_{i=1}^M\Big),
\end{align}
where ${\cal{N}}_c\in \mathbb{R}^{k\times 1}$ denotes the indices for $k$ nearest neighbors.
On top of this, the geometric patterns for each local neighborhood ${\cal{N}}_c$ are extracted by the delicate learnable operator, $\Phi(\cdot)$, and finally aggregated by max pooling, $\operatorname{MaxP}(\{\cdot\})$. We formulate it as
\begin{align}
    f_c^a = \operatorname{MaxP}\Big(\ \big\{\Phi(f_c, f_j)\big\}_{j \in {\cal{N}}_c}\Big).
\end{align}
The derived $\{p_c, f^a_c\}_{i=1}^\frac{M}{2}$ is then fed into the next network stage to progressively capture 3D geometries with enlarging receptive fields.

To verify the non-parametric efficacy independently from the learnable $\Phi(\cdot)$, we present Point-NN, a network purely constructed by these non-learnable basic components along with simple trigonometric functions for 3D coordinate encoding. Point-NN is composed of a non-parametric encoder $\operatorname{NPEnc}(\cdot)$ and a point-memory bank $\operatorname{PoM}(\cdot)$. Given an input point cloud $P$ for shape classification, the encoder summarizes its high-dimensional global feature $f_G$, and the bank produces the classification logits by similarity matching. We formulate it as
\begin{align}
\label{np}
    f_G = \operatorname{EncNP}(P); \ \  \mathrm{logits} = \operatorname{PoM}(f_G),
\end{align}
where $f_G \in \mathbb{R}^{1\times C_G}$ and the $\mathrm{logits}\in \mathbb{R}^{1 \times K}$ denote the predicted possibility for $K$ categories in the dataset.

\subsection{Non-Parametric Encoder}
\label{s3.2}
As shown in Figure~\ref{fig3}, the non-parametric encoder conducts initial embedding to transform the raw XYZ coordinates of $P$ into high-dimensional vectors, and progressively aggregates local patterns via the multi-stage hierarchy.

\paragraph{Raw-point Embedding.}
To achieve feature embedding without learnable layers, we refer to the positional encoding in Transformer~\cite{transformer} and extend it for non-parametric 3D encoding. For a raw point $i$ with $p_i=(x_i, y_i, z_i) \in \mathbb{R}^{1\times3}$, we utilize trigonometric functions to embed it into a $C_I$-dimensional vector,
\begin{align}
\operatorname{PosE}(p_i) = \operatorname{Concat}(f^{x}_i,\ f^{y}_i,\ f^{z}_i) \in \mathbb{R}^{1\times C_I},\label{equa5}
\end{align}
where $f^x_i,\ f^y_i,\ f^z_i \in \mathbb{R}^{1\times \frac{C_I}{3}}$ denote the embeddings of three axes, and $C_I$ denotes the initial feature dimension.
Taking $f^x_i$ as an example, for the channel index $m \in [0, \frac{C_I}{6}]$: 
\begin{align}
    f^x_i[2m] &= \operatorname{sine}\big(\alpha x_i/{\beta^{\frac{6m}{C_I}}}\big),\nonumber\\
    f^x_i[2m+1] &= \operatorname{cosine}\big({\alpha x_i}/{\beta^{\frac{6m}{C_I}}}\big), \label{equa6}
\end{align}
where $\alpha, \beta$ control the magnitude and wavelengths, respectively.
By the inherent nature of trigonometric functions, the relative position of two points can be revealed by a dot product between their embeddings, which captures fine-grained semantics of different local 3D structures.

\paragraph{Local Geometry Aggregation.}
Based on the embedding, we adopt a 4-stage network architecture to hierarchically aggregate spatial local features. After the ordinal FPS and $k$-NN illustrated in Section~\ref{s3.1}, we discard any learnable operators $\Phi(\cdot)$, and simply utilize trigonometric functions $\operatorname{PosE}(\cdot)$ to reveal the local patterns. In detail, for each center point $\{p_c, f_c\}$ and its neighborhood $\{p_{j}, f_j\}_{j\in{\cal N}_c}$, we aim to achieve three goals. 
\textbf{(1) \textit{Feature Expansion.}} As the network stage goes deeper, each point feature is assigned with larger receptive field and requires higher feature dimension to encode 3D semantics. We conduct such feature expansion by simply concatenating the neighbor feature $f_{j}$ with the center feature $f_c$ along the feature dimension,
\begin{align}
    f_{cj} = \operatorname{Concat}(f_c,\ f_{j}),\ \ \text{ for } j \in {\cal N}_c, \label{equa7}
\end{align}
where $f_{cj}$ denotes the expanded feature of each neighbor.
\textbf{(2) \textit{Geometry Extraction.}}
To indicate the spatial distribution of $k$ neighbors within the local region, we weigh each $f_{cj}$ by the relative positional encoding. We normalize their coordinates by the mean and standard deviation, denoted as $\{\Delta p_{j}\}_{j\in{\cal N}_c}$, and embed them via Eq.~\eqref{equa5}. Then, the $k$-neighbor features are weighted as
\begin{align}
\label{weigh}
    f^w_{cj} = \big(f_{cj} + \operatorname{PosE}(\Delta p_j)\big)  \odot \operatorname{PosE}(\Delta p_j),
\end{align}
where $\odot$ denotes element-wise multiplication. 
The feature dimension of $\operatorname{PosE}(\Delta p_j)$ is set the same as that of $f_{cj}$ in 4 stages, which are $2C_I$, $4C_I$, $8C_I$, and $16C_I$, due to feature expansion. 
In this way, the local geometry of the region, i.e., relative positional information of neighbor points $\operatorname{PosE}(\Delta p_j)$, can be implicitly encoded into the features without any learnable parameters. 
\textbf{(3) \textit{Feature Aggregation.}}
After weighing, we utilize both max and average pooling for local feature aggregation,\vspace{-0.1cm}
\begin{align}
    f_c^a= \operatorname{MaxP}\Big(\{f^w_{cj}\}_{j\in {\cal N}_c}\Big) + \operatorname{AveP}\Big(\{f^w_{cj}\}_{j \in {\cal N}_c}\Big), \label{equa9}
\end{align}
where $\operatorname{MaxP}(\{\cdot\}), \operatorname{AveP}(\{\cdot\})$ are permutation-invariant and summarize neighboring features from different aspects. Here, we obtain the local-aggregated centers $\{f^a_c, p_c\}_{c=1}^{\frac{M}{2}}$, which would be fed into the next stage of Point-NN. Finally, after all 4 stages, we apply the two pooling operations to integrate the features and acquire a global representation $f_G$ with $C_G$ feature dimension of the input point cloud.

\begin{figure}[t]
  \centering
\includegraphics[width=\textwidth]{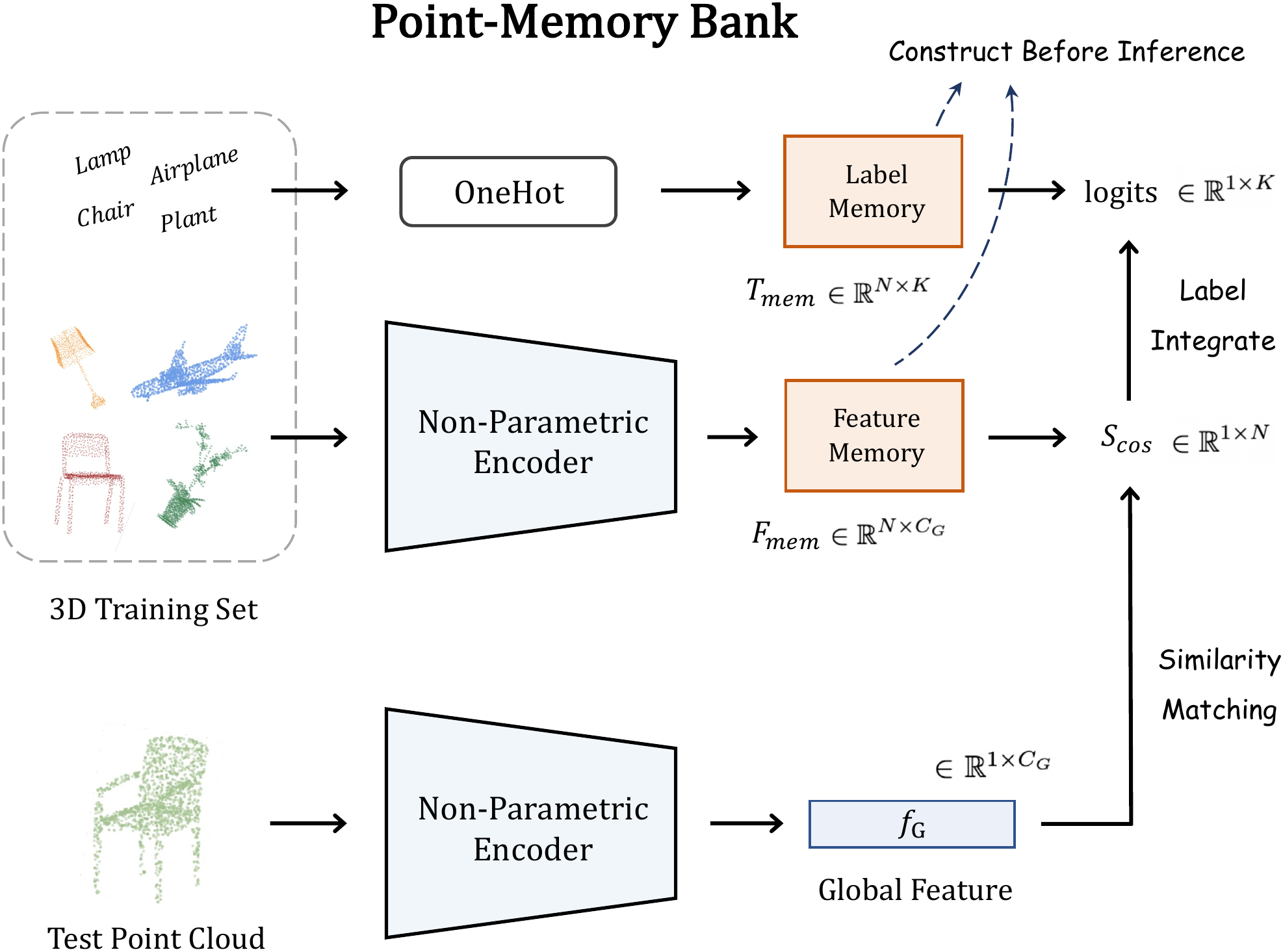}
\vspace{0.01cm}
   \caption{\textbf{Point-Memory Bank of Point-NN.} We construct the memory bank by caching training-set features via the non-parametric encoder. Then, the test point cloud is simply classified by similarity matching without training.}
    \label{fig4}
\vspace{-0.3cm}
\end{figure}

\subsection{Point-Memory Bank}
\label{s3.3}
Instead of using the traditional learnable classification head, our Point-NN adopts a point-memory bank to involve sufficient category knowledge from the 3D training set. As shown in Figure~\ref{fig4}, the bank is first constructed by the non-parametric encoder in a training-free manner, and then outputs the prediction by similarity matching during inference.

\vspace{-0.15cm}
\paragraph{Memory Construction.}
The point memory consists of a feature memory $F_{mem} $ and a label memory $T_{mem}$.
Taking the task of shape classification as an example, we suppose the given training set contains $N$ point clouds, $\{P_n\}_{n=1}^N$, of $K$ categories. Via the aforementioned non-parametric encoder, we encode all $N$ global features and convert their ground-truth labels $\{t_n\}_{n=1}^N$ as one-hot encoding. We then cache them as two matrices by concatenating along the inter-sample dimension as\vspace{-0.2cm}
\begin{align}
\label{prior}
    F_{mem} &= \operatorname{Concat}\Big(\big\{\operatorname{EncNP}(P_n)\big\}_{n=1}^N\Big),\vspace{-0.2cm}\\
    T_{mem} &= \operatorname{Concat}\Big(\big\{\operatorname{OneHot}(t_n)\big\}_{n=1}^N\Big),
\end{align}
where $F_{mem} \in \mathbb{R}^{N \times C_G}$ and $T_{mem} \in \mathbb{R}^{N \times K}$. Tagged by $T_{mem}$, the memory $F_{mem}$ can be regarded as the encoded category knowledge for the 3D training set. Features tagged with the same label unitedly describe the characteristics of the same category, and the inter-class discrimination can also be reflected by the embedding-space distances.

\vspace{-0.15cm}
\paragraph{Similarity-based Prediction.}
For a test point cloud, we also utilize the non-parametric encoder to extract its global feature as $f^t_{G} \in \mathbb{R}^{1 \times C_G}$, which is within the same embedding space as the feature memory $F_{mem}$. Then, the classification is simply accomplished by two matrix multiplications via the constructed bank. Firstly, we calculate the cosine similarity between the test feature and $F_{mem}$ by\vspace{-0.2cm}
\begin{align}
    S_{cos} = \frac{f_{G}^t F_{mem}^T}{\Vert f^t_{G}\Vert \cdot \Vert F_{mem}\Vert}\  \in \mathbb{R}^{1 \times N},
\end{align}
which denotes the semantic correlation of the test point cloud and $N$ training samples. Weighted by $S_{cos}$, we integrate the one-hot labels in the label memory $T_{mem}$ as
\begin{align}
    \mathrm{logits} = \varphi(S_{cos} T_{mem})\ \in \mathbb{R}^{1 \times K},
\end{align}
where $\varphi(x) = \exp(-\gamma(1 - x))$ serves as an activation function from~\cite{zhang2021tip}.
In $S_{cos}$, the more similar feature memory of higher score contributes more to the final classification logits, and vice versa. By such similarity-based label integration, the point-memory bank can adaptively discriminate different point cloud instances without any training.

\subsection{Other 3D Tasks}
\label{s3.4}

Except for shape classification, our Point-NN can also be extended for part segmentation and 3D object detection with tasks-specific modifications.

\begin{figure*}[t!]
  \centering
    \includegraphics[width=1\textwidth]{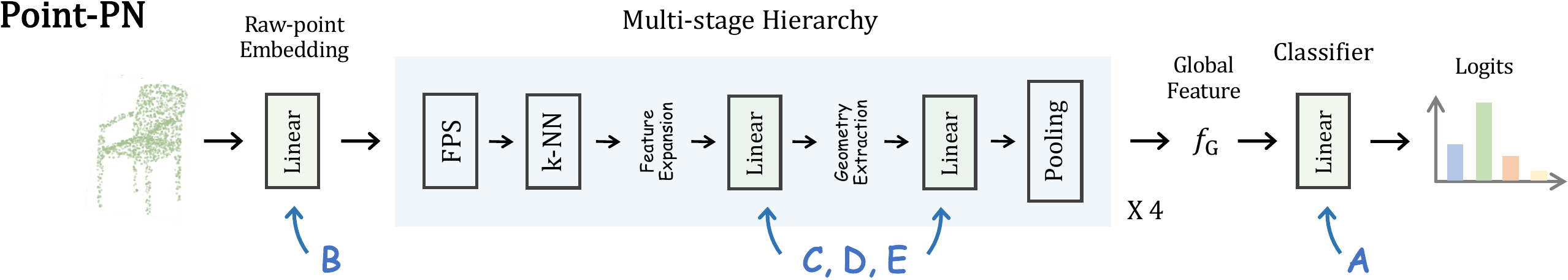}
   \caption{\textbf{The Pipeline of Point-PN.} Given the non-parametric framework of Point-NN, we simply construct the parametric derivative, Point-PN, by inserting linear layers into every stage of the encoder. Performance gain of using linear layers of A$\sim$E is shown in Table~\ref{t1}.}
    \label{fig5}
    \vspace{0.1cm}
\end{figure*}

\begin{figure*}[t!]
  \centering
    \includegraphics[width=0.94\textwidth]{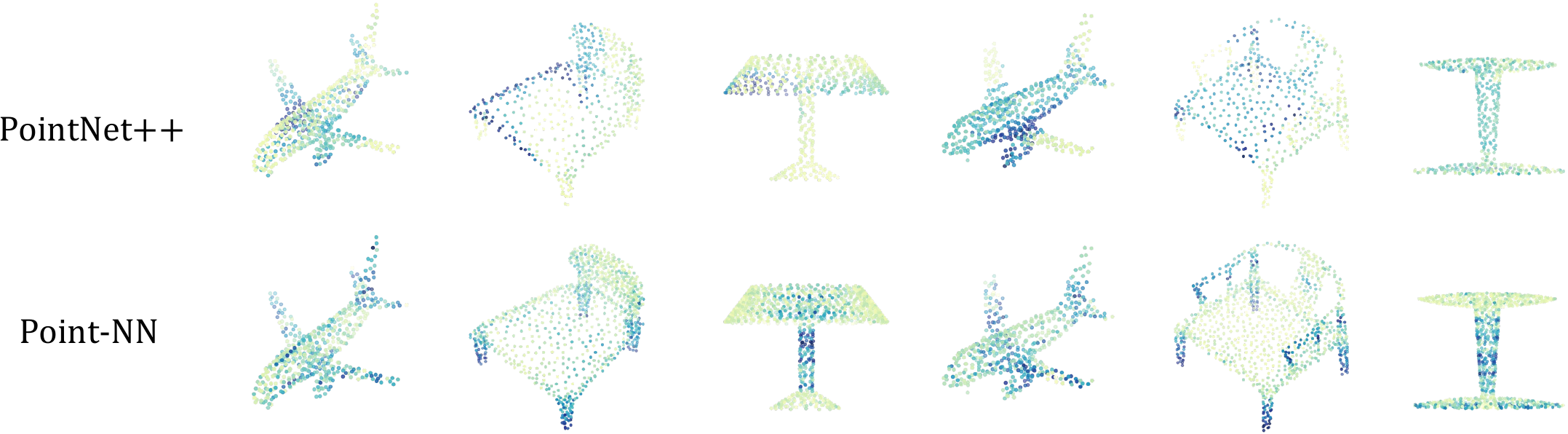}
   \caption{\textbf{Complementary Characteristics of Point-NN.} We visualize the point responses after the first network stage for the already trained PointNet++~\cite{qi2017pointnet++} and our Point-NN, where darker colors indicate higher responses. As shown, they focus on different spatial structures with complementary 3D patterns.}
    \label{fig7}
\end{figure*}

\paragraph{Part Segmentation.}
Other than extracting global features, the task of part segmentation requires to classify each input point. Therefore, we append a symmetric non-parametric decoder after the encoder, which progressively upsamples the encoded point features into the input point number. In every stage of the decoder, we reversely propagate the features of center points to their $k$ neighbors in a non-parametric manner. For the point-memory bank, we first apply the non-parametric encoder and decoder to extract all point-wise features of the training set. To save the GPU memory, we average the features of points with the same part label in an object, and only cache such aggregated part-wise features with part labels as $F_{mem}, T_{mem}$.\vspace{-0.2cm}

\paragraph{3D Object Detection.}
Given category-agnostic 3D proposals from a pre-trained 3D region proposal network~\cite{3detr,votenet}, Point-NN can be utilized as a non-parametric classification head for object detection. Similar to shape classification, we also adopt a pooling operation after the encoder to obtain global features of the detected objects. Differently, we sample the point cloud within each ground-truth 3D bounding box in the training set, and encode the object-wise features as the feature memory $F_{mem}$. Specifically, we do not normalize the point coordinates for each object as the pre-processing like other 3D tasks, which is to preserve the 3D positional information of objects in the original scene.

\section{Starting from Point-NN}

In this section, we introduce two promising applications for Point-NN, which fully unleash the potentials of non-parametric components for 3D point cloud analysis.

\begin{table}[t]
\centering

\begin{adjustbox}{width=\linewidth}
	\begin{tabular}{c c c cc c}
	\toprule
		\makecell*[c]{Method} &\makecell*[c]{Raw\\Embed.} &\makecell*[c]{Linear\\Layers} &\makecell*[c]{Classifier} &\makecell*[c]{Acc.\\(\%)}
		&\makecell*[c]{Param.}\\
		 \cmidrule(lr){1-1} \cmidrule(lr){2-4} \cmidrule(lr){5-5} \cmidrule(lr){6-6}
	    Point-NN &N &-  &N & 81.8 &0.0 M\\
	     A &N &-  &P & 90.3 &0.3 M\\
	     B &P &-  &P & 90.8 &0.3 M\\
	     C &P &0+1  &P & 93.4 &0.5 M\\
            D &P &1+1  &P & 93.2 &0.8 M\\
          E &P &2+2  &P & 92.9 &0.7 M\\
          \bf Point-PN &\bf P &\bf 1+2  &\bf P & \bf 93.8 &\bf 0.8 M\\
	\bottomrule
	\end{tabular}
\end{adjustbox}
\caption{\textbf{Step-by-step Construction of Point-PN} on ModelNet40~\cite{modelnet40}. `N' or `P' denotes the non-parametric modules or parametric linear layers. `Linear Layers' denotes the number of linear layers inserted into each network stage. Note that we adopt bottleneck layers with a ratio 0.5 for `2' in `Linear Layers'}
\label{t1}
\end{table}

\begin{table}[t]
\centering
\caption{\textbf{Shape Classification on the Real-world ScanObjectNN~\cite{scanobjectnn}}. We report the accuracy (\%) on three official splits of ScanObjectNN: OBJ-BG, OBJ-ONLY and PB-T50-RS. Our Point-NN outperforms the fully trained 3DmFV as marked in \color{blue}{blue}.}
\begin{adjustbox}{width=0.98\linewidth}
	\begin{tabular}{lcccc}
	\toprule
		\makecell*[c]{Method} &Split 1 &Split 2 &Split 3 &Param.\\
		\cmidrule(lr){1-1} \cmidrule(lr){2-2} \cmidrule(lr){3-3} \cmidrule(lr){4-4} \cmidrule(lr){5-5}
		3DmFV~\cite{3dmfv} &\color{blue}{68.2} &\color{blue}{73.8} &\color{blue}{63.0} &-\\
	    PointNet~\cite{qi2017pointnet}  &73.3 &79.2 &68.2 &3.5 M\\
	    SpiderCNN~\cite{xu2018spidercnn}  &77.1 &79.5 &73.7 &-\\
	    PointNet++~\cite{qi2017pointnet++} &82.3 &84.3 &77.9 &1.7 M\\
	    DGCNN~\cite{dgcnn} &82.8 &86.2 &78.1 &1.8 M\\
	    PointCNN~\cite{li2018pointcnn} &86.1 &85.5 &78.5 &-\\
	    DRNet~\cite{drnet} &- &- &80.3 &-\\
	    GBNet~\cite{gbnet} &- &-  &80.5 &8.4 M\\
	    SimpleView~\cite{simpleview} &- &-  &80.5 &-\\
	    PointMLP~\cite{pointmlp} &- &-  &85.2 &12.6 M\\
	    \cmidrule(lr){1-5}
	    \rowcolor{green!5}Point-NN\vspace{0.05cm} &\color{blue}{71.1} &\color{blue}{74.9}  &\color{blue}{64.9} &0.0 M\\
	    Point-PN &\textbf{\bf91.0} &\textbf{\bf90.2}  &\textbf{\bf87.1} &\bf0.8 M\\
	\bottomrule
	\end{tabular}
\end{adjustbox}
\vspace*{1pt}
\label{t2}
\end{table}

\subsection{As Architectural Frameworks}

Point-NN can be extended into learnable parametric networks (Point-PN) without adding complicated operators or too many parameters.
As shown in Figure~\ref{fig5} and Table~\ref{t1}, on top of Point-NN, we first replace the point-memory bank with a conventional learnable classifier. This lightweight version achieves 90.3\% classification accuracy on ModelNet40~\cite{modelnet40} with only 0.3M parameters (A). 
Then, we upgrade the raw-point embedding into parametric linear layers, which improves the performance to 90.8\% (B). To better extract multi-scale hierarchy, we append simple linear layers into every stage of the encoder. For each stage, two learnable linear layers are inserted right before or after the \textit{Geometry Extraction} step for capturing higher-level spatial patterns (C, D, E). 
We observe Point-PN attains the competitive 93.8\% accuracy with 0.8M parameters. This final version only contains trigonometric functions for geometry extraction and simple linear layers for feature transformation.
This demonstrates that, compared to existing advanced operators or scaled-up parameters, we can alternatively start from a non-parametric framework, i.e., Point-NN, to obtain a powerful and efficient 3D model.  

\subsection{As Plug-and-play Modules}

Considering the training-free characteristic, we propose to regard Point-NN as an inference-time enhancement module, which can boost already trained 3D models without extra re-training. For shape classification, we directly fuse the prediction by linear interpolation, namely, adding the classification logits of Point-NN and off-the-shelf models element-wisely. This design produces the ensemble for two types of knowledge: the low-level structural signals from Point-NN, and the high-level semantics from the trained networks. As visualized in Figure~\ref{fig7}, the extracted point cloud features by Point-NN produce high response values around the sharp 3D structures, e.g., the airplane's wingtips, chair's legs, and lamp poles. In contrast, the trained PointNet++ focuses more on 3D structures with rich semantics, e.g., airplane's main body, chair's bottoms, and lampshades.


\vspace{0.2cm}
\section{Experiments}
\vspace{0.2cm}

We conduct extensive experiments to evaluate the efficacy of Point-NN (Sec.~\ref{s4.1}), and its two extensions as Point-PN (Sec.~\ref{s4.2}) and plug-and-play modules (Sec.~\ref{s4.3}).

\begin{table}[t]
\centering
\caption{\textbf{Shape Classification on Synthetic ModelNet40~\cite{modelnet40}}. All compared methods take 1,024 points as input. Train Time and Test Speed (samples/second) are tested on one RTX 3090 GPU. We report the accuracy without the voting strategy.\vspace{0.3cm}}
\begin{adjustbox}{width=0.997\linewidth}
	\begin{tabular}{lcccc}
	\toprule
		\makecell*[c]{Method} &Acc. (\%) &Param. &\makecell*[c]{Train\\Time} &\makecell*[c]{Test\\Speed}\\
		\cmidrule(lr){1-1} \cmidrule(lr){2-2} \cmidrule(lr){3-3} \cmidrule(lr){4-4} \cmidrule(lr){5-5}
	    PointNet~\cite{qi2017pointnet}  &89.2 &3.5 M &-&-\\
	    PointNet++~\cite{qi2017pointnet++} &90.7 &1.7 M  &3.4~h &521\\
	    DGCNN~\cite{dgcnn} &92.9 &1.8 M  &\textbf{2.4~h}&617\\
	    RS-CNN~\cite{rscnn} &92.9 &1.3 M  &-&-\\
	    DensePoint~\cite{densepoint} &93.2 &- &-&-\\
	    PCT~\cite{guo2021pct} &93.2 &-  &-&-\\
	    GBNet~\cite{gbnet} &93.8 &8.4 M  &-&189\\
	    CurveNet~\cite{curvenet} &93.8 &2.0~M  &6.7~h&25\\
	    PointMLP~\cite{pointmlp} &\textbf{94.1} &12.6 M  &14.4~h &189\\
	    \cmidrule(lr){1-5}
	    \rowcolor{green!5} Point-NN\vspace{0.05cm} &{81.8} &{0.0 M}  &0 &275\\
	    Point-PN &93.8 &\textbf{0.8 M}  &3.9~h&\textbf{1176}\\
	\bottomrule
	\end{tabular}
\end{adjustbox}
\label{t3}
\end{table}

\begin{table}[t]
\centering
\caption{\textbf{Part Segmentation on ShapeNetPart~\cite{shapenetpart}}. All compared methods take 2,048 points as input, and are evaluated by mean IoU scores (\%) across instances.}
\begin{adjustbox}{width=0.94\linewidth}
	\begin{tabular}{lcccc}
	\toprule
		\makecell*[c]{Method} &\makecell*[c]{Inst.\\mIoU} &Param. &\makecell*[c]{Train\\Time} &\makecell*[c]{Test\\Speed}\\
		\cmidrule(lr){1-1} \cmidrule(lr){2-2} \cmidrule(lr){3-3} \cmidrule(lr){4-4} \cmidrule(lr){5-5}
	    PointNet~\cite{qi2017pointnet}  &83.7 &8.3~M &- &-\\
	    PointNet++~\cite{qi2017pointnet++} &85.1 &\textbf{1.8~M}  &\textbf{26.5~h} &45\\
	    PAConv~\cite{guo2021pct} &86.0 &-  &- &-\\
	    PointMLP~\cite{pointmlp} &86.1 &16.8~M  &47.1~h &119\\
	    CurveNet~\cite{curvenet} &\textbf{86.6} &5.5 M  &56.9~h &22\\
	    \cmidrule(lr){1-5}
	    \rowcolor{green!5}{Point-NN}\vspace{0.05cm} &{70.4} &{0.0 M}  &{0}&51\\
	    Point-PN &\textbf{86.6} &3.9 M  &29.0~h &\textbf{131}\\
	\bottomrule
	\end{tabular}
\end{adjustbox}
\vspace*{1pt}
\label{t5}
\end{table}

\vspace{0.2cm}

\subsection{Dataset}
\label{s4.0}
For \textbf{shape classification}, we report the performance on two benchmarks: the synthetic ModelNet40~\cite{modelnet40} with 40 categories, and real-world ScanObjectNN~\cite{scanobjectnn} with 15 categories. Considering the background and data augmentation, ScanObjectNN is officially split into three subsets: OBJ-BG, OBJ-ONLY, and PB-T50-RS. 
For \textbf{few-shot classification}, we evaluate on the few-shot subset of ModelNet40 with four different settings, 5-way 10-shot, 5-way 20-shot, 10-way 10-shot and 10-way 20-shot.
For \textbf{part segmentation}, we adopt ShapeNetPart~\cite{shapenetpart} dataset with synthesized 3D shapes of 50 part categories.
For \textbf{3D object detection}, we conduct the experiments on ScanNetV2~\cite{ScanNetV2} with axis-aligned bounding boxes in 3D scenes. We refer to Supplementary Material for implementation details. Note that, we report all results \textbf{without the voting strategy}.

\vspace{0.2cm}
\subsection{Point-NN}
\label{s4.1}
\vspace{0.2cm}

\paragraph{Shape Classification.}
The results of Point-NN are reported in the green rows of Table~\ref{t2} and \ref{t3}, respectively for the two datasets. As shown, Point-NN attains favorable classification accuracy for both real-world and synthetic point clouds, indicating our effectiveness and generalizability. Surprisingly, Point-NN even surpasses the fully trained 3DmFV~\cite{3dmfv} by +2.9\%, +1.1\%, +1.9\% on the three splits of ScanObjectNN~\cite{scanobjectnn}. By this, we demonstrate that, without any parameters or training, the non-parametric network components with simple trigonometric functions can achieve satisfactory 3D point cloud recognition.
\vspace{-0.1cm}

\begin{table}[t]
\centering
\caption{\textbf{Few-shot Classification on ModelNet40~\cite{modelnet40}}. We calculate the average accuracy ($\%$) over 10 independent runs. The reported results of existing methods are taken from ~\cite{sharma2020self}.}
\begin{adjustbox}{width=\linewidth}
	\begin{tabular}{lc c c c c}
	\toprule
		\makecell*[c]{\multirow{2}*{Method}} &\multicolumn{2}{c}{5-way} &\multicolumn{2}{c}{10-way}\\
		 \cmidrule(lr){2-3} \cmidrule(lr){4-5}
		 &10-shot &20-shot &10-shot &20-shot\\
		 \cmidrule(lr){1-1} \cmidrule(lr){2-5}
		DGCNN~\cite{dgcnn} &31.6 &40.8  &19.9 &16.9\\
            FoldingNet~\cite{yang2018foldingnet} &33.4 &35.8 &18.6 &15.4\\
	    PointNet++~\cite{qi2017pointnet++}  &38.5 &42.4 &23.0 &18.8\\
	    PointNet~\cite{qi2017pointnet} &52.0 &57.8  &46.6 &35.2\\
	    3D-GAN~\cite{wu2016learning} &55.8 &65.8  &40.3 &48.4\\
	    PointCNN~\cite{li2018pointcnn} &65.4 &68.6  &46.6 &50.0\\
	    \cmidrule(lr){1-5}
	    \rowcolor{green!5}Point-NN\vspace{0.05cm} &88.8 &90.9  &79.9 &84.9\\
	    \textit{Improvements} &\textcolor{blue}{+23.4} &\textcolor{blue}{+22.3} &\textcolor{blue}{+33.3} &\textcolor{blue}{+34.9}\\
	\bottomrule
	\end{tabular}
\end{adjustbox}
\label{t4}
\end{table}

\paragraph{Few-shot Classification.}
As shown in Table~\ref{t4}, compared to existing trained models, our Point-NN exerts leading few-shot performance and exceeds the second-best PointCNN~\cite{li2018pointcnn} by significant margins, e.g., +34.9\% for the 10-way 20-shot setting. Given limited training samples, traditional networks with learnable parameters severely suffer from the over-fitting issue, while our Point-NN inherently tackles it by the training-free manner. This indicates our superior capacity for 3D recognition in low-data regimes.

\paragraph{Part Segmentation.}\vspace{-0.1cm}
Equipped with a non-parametric decoder illustrated in Sec.~\ref{s3.4}, we extend Point-NN for part segmentation and report the mIoU scores over instances in the green row of Table~\ref{t5}.
The 70.4\% mIoU reveals that our non-parametric network can also produce well-performed point-wise features, and capture the discriminative characteristics for fine-grained spatial understanding.

\vspace{-0.1cm}
\paragraph{3D Object Detection.}
Regarding Point-NN as a non-parametric classification head, we utilize two popular 3D detectors, VoteNet~\cite{votenet} and 3DETR-m~\cite{3detr}, to provide category-agnostic 3D region proposals. In Table~\ref{t6}, we compare the performance of Point-NN with and without the normalization for object-wise point coordinates (Sec.~\ref{s3.4}). As shown, processing the point coordinates without normalization can largely enhance the AP scores of Point-NN, which preserves more positional cues of objects' 3D locations in the original scene. Also, we observe slight AR score improvement over the region proposal networks, since the classification logits of Point-NN can affect the 3D NMS post-processing to remove the false positive boxes.

\vspace{-0.1cm}
\paragraph{Ablation Study.}
In Table~\ref{tab1}, we investigate the designs of Point-NN's non-parametric encoder. We first explore the non-parametric embedding used as PosE, where the trigonometric functions (`Our Sin/cos') from Transformers~\cite{transformer} perform the best. Then, we experiment with different approaches in \textit{Geometry Extraction} to weigh the $k$-neighbor features, and observe utilizing `Add+Multiply' can fully reveal the local geometry.
In Table~\ref{tab2} for Point-NN's point-memory bank, we verify that the cosine similarity better exerts the discrimination capacity of the encoder among other distance metrics. In addition, compared to traditional machine learning algorithms, our memory bank can benefit from the non-learnable similarity-based label integration, and exhibit superior classification accuracy.

\begin{table}[t]
\centering
\caption{\textbf{3D Object Detection on ScanNetV2~\cite{ScanNetV2}}. We report the mean Average Precision (\%) and mean Average Recall (\%) with 0.25 and 0.5 IoU thresholds. `nor' denotes to normalize the point coordinates for each object proposal.}
\begin{adjustbox}{width=0.97\linewidth}
	\begin{tabular}{lc c c c c}
	\toprule
		 \makecell*[c]{Method} &\ \ AP$_{25}$\ &\ AP$_{50}$\ \ &\ \ AR$_{25}$\ &\ AR$_{50}$\ \ \\
		 \cmidrule(lr){1-1} \cmidrule(lr){2-3} \cmidrule(lr){4-5}
	    VoteNet~\cite{votenet}  &57.8 &33.5 &80.9 &51.3\\
	    Point-NN\vspace{0.1cm} &4.5 &3.3 &80.9 &51.3\\
	    \rowcolor{green!5}Point-NN \textbf{w/o nor.} &23.1 &16.0 &90.0  &51.3\\
	    \cmidrule(lr){1-5}
	    3DETR-m~\cite{3detr} &64.6  &46.4 &77.2 &59.2\\
	    Point-NN\vspace{0.05cm} &7.7 &5.7 &77.2  &59.2\\
	    \rowcolor{green!5}Point-NN \textbf{w/o nor.} &33.3 &24.7 &77.4  &59.3\\
	\bottomrule
	\end{tabular}
\end{adjustbox}
\vspace*{2pt}
\label{t6}
\end{table}

\subsection{Point-PN}
\label{s4.2}
\vspace{0.1cm}

\paragraph{Shape Classification.}
As shown in Table~\ref{t2} and \ref{t3}, constructed from Point-NN, the derived Point-PN achieves competitive results for both datasets. Compared to the large-scale PointMLP~\cite{pointmlp} with stacked MLPs of 12.6M parameters, Point-PN only contains simple linear layers and surpasses it by +1.9\% accuracy on ScanObjectNN~\cite{scanobjectnn} with 16$\times$ fewer parameters and 6$\times$ faster inference speed. With simple trigonometric functions, Point-PN attains comparable performance to CurveNet~\cite{curvenet} with complicated curve-based grouping on ModelNet40~\cite{modelnet40}, while contains 2.5$\times$ fewer parameters and 6$\times$ faster inference speed. The superior classification accuracy of Point-PN fully demonstrates the significance of a powerful non-parametric framework.

\vspace{-0.15cm}
\paragraph{Part Segmentation.}
For point-wise segmentation task in Table~\ref{t5}, Point-PN also achieves competitive performance, i.e., 86.6\% mIoU, among existing methods. Compared to CurveNet, Point-PN with simple local geometry aggregation can save 28h training time and inference 6$\times$ faster.

\begin{table}[t]
\tabcaption{\textbf{Ablation Study of Non-Parametric Encoder.} We ablate the non-parametric functions for point embedding, and experiment different weighing methods for local geometry extraction.}
\begin{minipage}[t]{0.457\linewidth}
\centering
\begin{adjustbox}{width=\linewidth}
	\begin{tabular}{cc}
	\toprule
		\makecell[c]{Embedding\\Function} &Acc (\%) \\
        \cmidrule(lr){1-2}
        w/o & 68.6\\
        NeRF's~\cite{mildenhall2021nerf} &70.1 \\
	  Fourier's~\cite{tancik2020fourier} &76.9  \\
        \textbf{\bf Our Sin/cos} &\textbf{\bf81.8} \\
	\bottomrule
	\end{tabular}
\end{adjustbox}
\end{minipage}\quad
\begin{minipage}[t!]{0.48\linewidth}
\centering
\begin{adjustbox}{width=\linewidth}
	\begin{tabular}{cc}
	\toprule
		\makecell[c]{Weighted\\ by PosE} &Acc (\%)\\
        \cmidrule(lr){1-2}
		w/o &77.8 \\
        Add &78.3 \\
        Multiply &80.4 \\
        \textbf{\bf Add+Multiply} &\textbf{\bf81.8} \\
	\bottomrule
	\end{tabular}
\end{adjustbox}
\end{minipage}
\label{tab1}
\vspace*{-2pt}
\end{table}

\vspace{1.7cm}
\begin{table}[t]
\centering
\vspace*{1pt}
\caption{\textbf{Plug-and-play for Shape Classification.} We report the gain (\%) on the PB-T50-RS split of ScanObjectNN.}
\begin{adjustbox}{width=\linewidth}
	\begin{tabular}{lc c c c c}
	\toprule
		\makecell*[c]{Method} &ScanObjNN &+NN &ModelNet40 &+NN\\
		 \cmidrule(lr){1-1} \cmidrule(lr){2-3} \cmidrule(lr){4-5}
            Point-NN &64.9 &- &81.8 &-\\
		PointNet &68.2 &\textcolor{blue}{+2.2} &89.7 &\textcolor{blue}{+0.4} \\
        PointNet++  &77.9 &\textcolor{blue}{+1.2} &92.6 &\textcolor{blue}{+0.5}\\
          PCT &-&-&93.2 &\textcolor{blue}{+0.2}\\
	    PointMLP &85.2 &\textcolor{blue}{+2.0}&94.1 &\textcolor{blue}{+0.3} \\
	\bottomrule
	\end{tabular}
\end{adjustbox}
\label{t8}
\vspace*{-2pt}
\end{table}

\vspace{-1.5cm}
\paragraph{Ablation Study.}
In Table~\ref{t1}, we present how to step-by-step construct Point-PN from the non-parametric Point-NN. As illustrated in Section~\ref{s4.1}, we insert linear layers before and after the \textit{Geometry Extraction} step. `C' of `0+1' denotes only appending one linear layer after the module, and `D' of `1+1' denotes inserting into both positions. The preceding layers pre-transform the point features to better reveal the local geometry, and the latter layers further parse the weighted features for high-level understanding.
We observe that Point-PN of `1+2' performs the best.


\vspace{0.1cm}
\subsection{Plug-and-play}
\vspace{0.1cm}
\label{s4.3}

\paragraph{Shape Classification.}
We evaluate the enhancement capacity of Point-NN on two classification datasets in Table~\ref{t8}. By simple inference-time ensemble, Point-NN effectively boosts existing methods with different margins. On the more challenging ScanObjectNN~\cite{scanobjectnn}, both PointNet~\cite{qi2017pointnet} and PointMLP~\cite{pointmlp} are improved by +2.0\% accuracy. This well indicates the effectiveness of complementary knowledge provided by Point-NN.

\begin{table}[t]
\tabcaption{\textbf{Ablation Study of Point-Memory Bank.} We compare the similarity metrics and machine learning algorithms. `GBoost' and `Dec. Tree' denote Gradient Boosting and Decision Tree.}
\begin{minipage}[t!]{0.405\linewidth}
\centering
\begin{adjustbox}{width=\linewidth}
	\begin{tabular}{cc}
	\toprule
		\makecell[c]{Similarity\\Metric} &Acc (\%) \\
        \cmidrule(lr){1-2}
        Chebyshev &65.9 \\
        Euclidean &79.8 \\
        Manhattan &80.9 \\
        \textbf{\bf Cosine} &\textbf{\bf81.8} \\
	\bottomrule
	\end{tabular}
\end{adjustbox}
\end{minipage}\quad
\begin{minipage}[t!]{0.475\linewidth}
\centering
\begin{adjustbox}{width=\linewidth}
	\begin{tabular}{cc}
	\toprule
		\makecell[c]{Traditional\\Classifier} &Acc (\%)\\
        \cmidrule(lr){1-2}
        Dec. Tree~\cite{safavian1991survey} &57.8 \\
        GBoost~\cite{friedman2001greedy} &63.9 \\
        SVM~\cite{cristianini2000introduction} &79.9 \\
        \textbf{\bf Memory Bank} &\textbf{\bf81.8} \\
	\bottomrule
	\end{tabular}
\end{adjustbox}
\end{minipage}
\vspace*{-1pt}
\label{tab2}
\end{table}

\begin{table}[t]
\vspace*{4pt}
\tabcaption{\textbf{Plug-and-play for Segmentation and Detection.} We report the gain (\%) on ShapeNetPart and ScanNetV2, respectively.}
\vspace*{4pt}
\begin{minipage}[t!]{0.35\linewidth}
\centering
\begin{adjustbox}{width=\linewidth}
	\begin{tabular}{lcc}
	\toprule
		\makecell*[c]{Method} &mIoU \\
		 \cmidrule(lr){1-1} \cmidrule(lr){2-2}
	    DGCNN &85.16  \\
	    +NN &\textcolor{blue}{+0.16} \\
     \cmidrule(lr){1-2}
	    CurveNet &86.58 \\
	    +NN &\textcolor{blue}{+0.07} \\
	\bottomrule
	\end{tabular}
\end{adjustbox}
\end{minipage}\quad
\begin{minipage}[t!]{0.53\linewidth}
\centering
\begin{adjustbox}{width=\linewidth}
	\begin{tabular}{lccc}
	\toprule
		\makecell*[c]{Method} &AP$_{25}$ &AR$_{25}$\\
		 \cmidrule(lr){1-1} \cmidrule(lr){2-3}
	    VoteNet &57.84 &80.92 \\
	    +NN &\textcolor{blue}{+1.28} &\textcolor{blue}{+5.31}\\
	    \cmidrule(lr){1-3}
	    3DETR-m &64.60 &77.22\\
	    +NN &\textcolor{blue}{+1.02} &\textcolor{blue}{+11.05} \\
	\bottomrule
	\end{tabular}
\end{adjustbox}
\end{minipage}
\vspace*{-1pt}
\label{t10}
\end{table}

\vspace{-0.15cm}
\paragraph{Segmentation and Detection.}
In Table~\ref{t10}, we present the plug-and-play performance of Point-NN on part segmentation and 3D object detection tasks. As the segmentation scores have long been saturated on the ShapeNetPart~\cite{shapenetpart} benchmark, the boost of +0.1\% mIoU for state-of-the-art CurveNet~\cite{curvenet} is still noteworthy. For detection, Point-NN significantly enhances 3DETR-m~\cite{3detr} by +1.02\% AP$_{25}$ and +11.05\% AR$_{25}$. By fusing complementary knowledge to the trained classifier, the 3D detector can better judge whether the candidate boxes include objects and correctly remove the false positive ones.

\begin{figure*}[t!]
  \centering
    \includegraphics[width=1\textwidth]{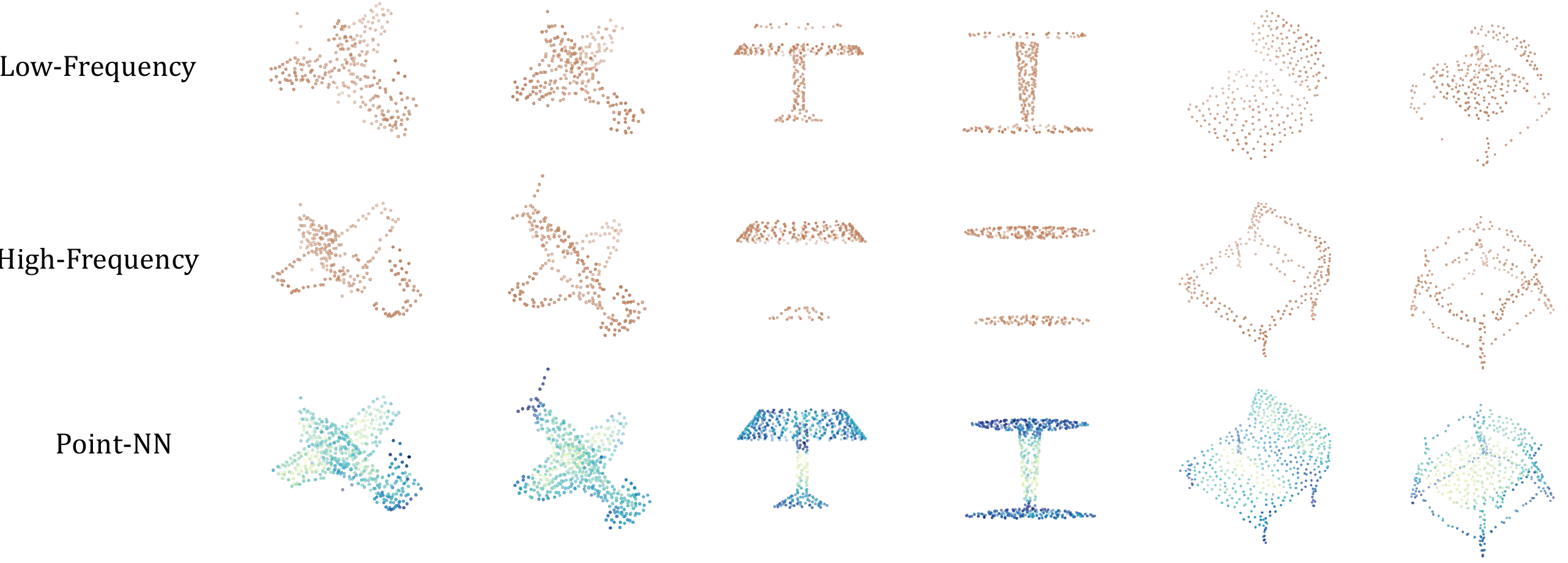}
   \caption{\textbf{Why Do Trigonometric Functions Work?} For an input point cloud, we visualize its low-frequency and high-frequency geometries referring to~\cite{xu2021learning}, and compare with the feature responses after the first network stage of Point-NN, where darker colors indicate higher responses. As shown, Point-NN can focus on the high-frequency 3D structures with sharp variations of the point cloud.}
    \label{supp_f11}
\end{figure*}

\section{Discussion}
\label{s4}

\subsection{Why Do Trigonometric Functions Work?}

We leverage the trigonometric function to conduct non-parametric raw-point embedding and geometry extraction. It can reveal the 3D spatial patterns benefited from the following three properties.

\paragraph{Capturing High-frequency 3D Structures.}
As discussed in Tancik et al.~\cite{tancik2020fourier}, transforming low-dimensional input by sinusoidal mapping helps MLPs to learn the high-frequency content during training. Similarly to our non-parametric encoding, Point-NN utilizes trigonometric functions to capture the high-frequency spatial structures of 3D point clouds, and then recognize them from these distinctive characteristics by the point-memory bank. In Figure~\ref{supp_f11}, we visualize the low-frequency (Top) and high-frequency (Middle) geometry of the input point cloud, and compare them with the feature responses of Point-NN (Bottom). The high-frequency geometries denotes the spatial regions of edges, corners, and other fine-grained details, where the local 3D coordinates vary dramatically, while the low-frequency structure normally includes some flat and smooth object surfaces with gentle variations. As shown, aided by trigonometric functions, our Point-NN can concentrate well on these high-frequency 3D patterns.

\begin{figure*}[t!]
  \centering
    \includegraphics[width=0.94\textwidth]{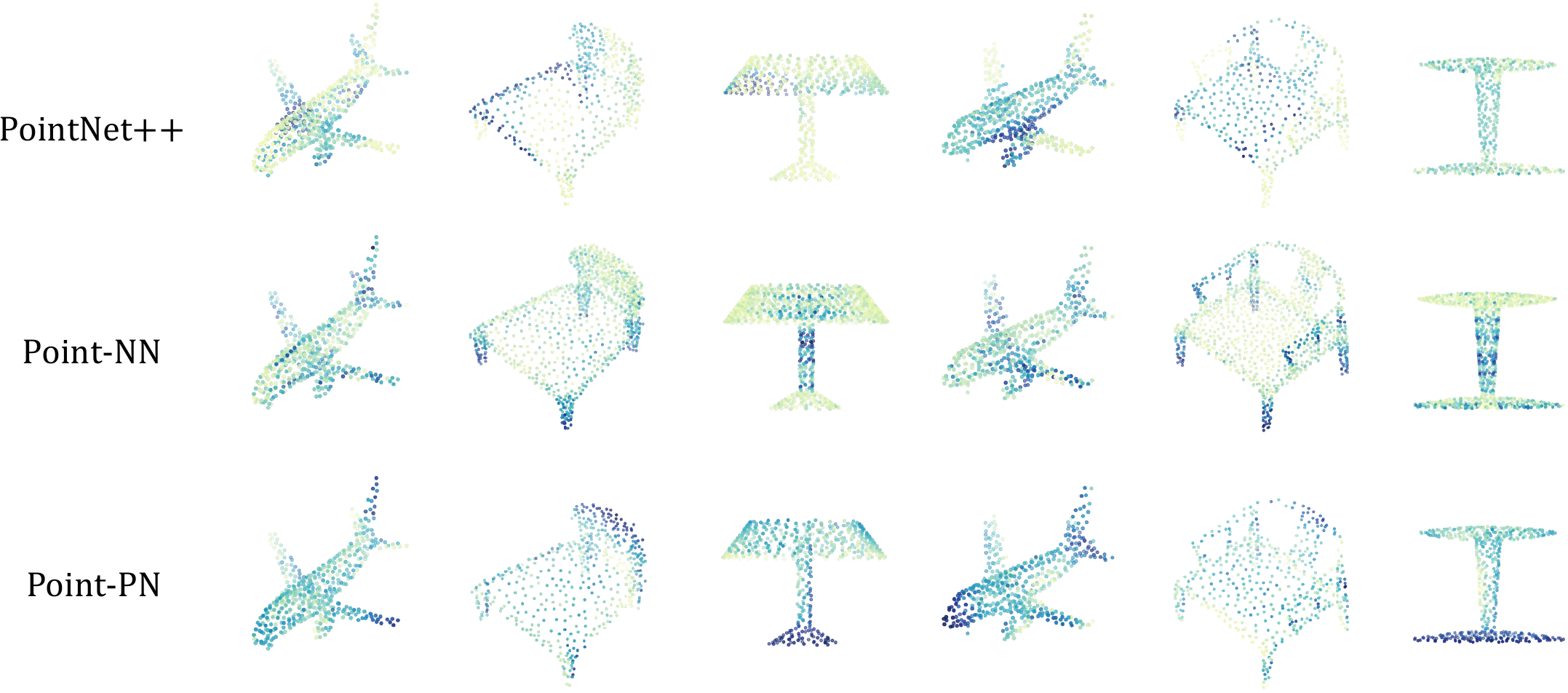}
   \caption{\textbf{Can Point-NN Improve Point-PN by Plug-and-play?} We visualize the feature responses for Point-NN, the trained PointNet++~\cite{qi2017pointnet++} and Point-PN, where darker colors indicate higher responses. As shown, Point-PN captures similar 3D patterns to Point-NN, which harms their complementarity.}
    \label{supp_f2}
\end{figure*}

\paragraph{Encoding Absolute and Relative Positions.}
Benefited from the nature of sinusoid, the trigonometric functions can not only represent the absolute position in the embedding space, but also implicitly encode the relative positional information between two 3D points. For two points, $p_i=(x_i, y_i, z_i)$ and $p_j=(x_j, y_j, z_j)$, we first obtain their $C$-dimensional embeddings referring to Equation \eqref{equa5}$\sim$\eqref{equa7} in the main paper, formulated as
\begin{align}
\operatorname{PosE}(p_i) &= \operatorname{Concat}(f^{x}_i,\ f^{y}_i,\ f^{z}_i),\\
\operatorname{PosE}(p_j) &= \operatorname{Concat}(f^{x}_j,\ f^{y}_j,\ f^{z}_j),
\end{align}
where $\operatorname{PosE}(\cdot)$ denotes the positional encoding by trigonometric functions, and  $f^x_{i/j},\ f^y_{i/j},\ f^z_{i/j} \in \mathbb{R}^{1\times \frac{C}{3}}$ denote the embeddings of three axes.
Then, their spatial relative relation can be revealed by the dot production between the two embeddings, formulated as
\begin{align}
f^{x}_i f^{xT}_j + f^{y}_i f^{yT}_j + f^{z}_i f^{zT}_j = \operatorname{PosE}(p_i) \operatorname{PosE}(p_j)^T.\nonumber
\end{align}
Taking the x axis as an example,
\begin{align}
\sum_{m=0}^{\frac{C}{6}-1}\mathrm{cosine}\big(\alpha (x_i - x_j)/{\beta^{\frac{6m}{C}}}\big) = f^{x}_i f^{xT}_j,
\end{align}
which indicates the relative x-axis distance of two points, in a similar way to the other two axes. Therefore, the trigonometric function is capable of encoding both absolute and relative 3D positional information for point cloud analysis.


\paragraph{Local Geometry Extraction.}
In Equation \eqref{equa9} of the main paper, we weigh each neighbor feature $f_j$ within the local region by the relative positional embedding, $\operatorname{PosE}(\Delta p_j)$, formulated as
\begin{align}
\label{weigh2}
    f^w_{cj} &= \big(f_{cj} + \operatorname{PosE}(\Delta p_j)\big) \odot \operatorname{PosE}(\Delta p_j).
\end{align}
The weighing is conducted sequentially by element-wise addition and multiplication. 
Firstly, the addition is to complement $f_{cj}$ with higher frequency information. Due to feature expansion, the output dimensions of $\operatorname{PosE}(\Delta p_j)$ of 4 stages are respectively $2C_I$, $4C_I$, $8C_I$, and $16C_I$. As the embedding frequency depends on feature dimension referring to Equation \eqref{equa6} of the main paper, the embeddings at higher stages obtain higher frequencies. Taking the first stage as an example, $\operatorname{PosE}(\Delta p_j)$ is $2C_I$-dimensional, but $f_{cj}$ is a concatenation of two $C_I$-dimensional vectors, which makes their embedding frequencies inconsistent. Therefore, we adopt addition to endow $f_{cj}$ with the frequency corresponding to $2C_I$ dimension.
Then, the second-step multiplication weighs the magnitude of each element in $f_{cj}$ by its relative positional information. This determines the importance of different neighbor points in the subsequent pooling operations, and the final aggregated features of the local neighborhood. In this way, Point-NN can effectively embed local 3D patterns via $\operatorname{PosE}(\cdot)$ without any learnable operators.

\begin{table}[t]
\centering
\caption{\textbf{Can Point-NN Improve Point-PN by Plug-and-play?} We report the accuracy (\%) on the PB-T50-RS split of ScanObjectNN~\cite{scanobjectnn}, ModelNet40~\cite{modelnet40}, and ShapeNetPart~\cite{shapenetpart}.}
\begin{adjustbox}{width=0.92\linewidth}
	\begin{tabular}{cccc}
	\toprule
	\makecell*[c]{Benchmark} &ScanObjectNN &ModelNet40 &ShapeNetPart\\
        \cmidrule(lr){1-1} \cmidrule(lr){2-4}
        \specialrule{0em}{1pt}{1pt}
		 Point-PN &87.1 &93.8  &86.6 \\
		 \specialrule{0em}{1pt}{1pt}
         +NN &\textcolor{blue}{+0.1} &\textcolor{blue}{+0.2} &\textcolor{blue}{+0.0}\\ 
		 \specialrule{0em}{1pt}{1pt}
	\bottomrule
	\end{tabular}
\end{adjustbox}
\label{supp_t8}
\end{table}

\vspace{0.2cm}
\subsection{Can Point-NN Improve Point-PN?}
Point-NN can provide complementary geometric knowledge and serve as a plug-and-play module to boost existing learnable 3D models. Although Point-PN is also a learnable 3D network, the enhanced performance brought by Point-NN is marginal as reported in Table~\ref{supp_t8}. By visualizing feature responses in Figure~\ref{supp_f2}, we observe that the complementarity between Point-NN and Point-PN is much weaker than that between Point-NN and PointNet++~\cite{qi2017pointnet++}. This is because the non-parametric framework of Point-PN is mostly inherited from Point-NN, also capturing high-frequency 3D geometries via trigonometric functions. Therefore, the learnable Point-PN extracts similar 3D patterns to Point-NN, which harms its plug-and-play capacity.

\begin{table}[t]
\centering
\caption{\textbf{Comparison of Training-free Methods in 3D.} We report their performance without training on ModelNet40~\cite{modelnet40}.}
\vspace{-0.1cm}
\begin{adjustbox}{width=\linewidth}
	\begin{tabular}{lcccc}
	\toprule
	\makecell*[c]{Method} &\makecell*[c]{Pre-train\\in 2D} &\makecell*[c]{Pre-train\\in 3D} &3D Data &Acc. (\%)\\
        \cmidrule(lr){1-1} \cmidrule(lr){2-5}
        \specialrule{0em}{1pt}{1pt}
		 PointCLIP~\cite{zhang2022pointclip} &\checkmark &-  &- &20.2 \\
        CALIP~\cite{guo2022calip} &\checkmark &-  &- &21.5 \\
        CLIP2Point~\cite{huang2022clip2point} &\checkmark &\checkmark  &\checkmark &49.4\\
        ULIP~\cite{xue2022ulip} &\checkmark &\checkmark  &\checkmark &60.4\\
        PointCLIP V2~\cite{zhu2022pointclip} &\checkmark &- &-  &64.2 \\
		 \specialrule{0em}{1pt}{1pt}
         \textbf{Point-NN} &- &- &\checkmark &\textbf{81.8}\\ 
		 \specialrule{0em}{1pt}{1pt}
	\bottomrule
	\end{tabular}
\end{adjustbox}
\label{supp_t9}
\end{table}

\vspace{0.2cm}
\subsection{Training-free Methods in 3D}

Our Point-NN conducts no training, but requires 3D training data to construct the point-memory bank. Inspired by the transfer learning in 2D and language, some recent works~\cite{zhu2022pointclip,zhang2022pointclip,huang2022clip2point,guo2022calip,xue2022ulip} adapt the pre-trained models from other modalities, e.g., CLIP~\cite{radford2021learning}, to 3D domains in a zero-shot manner. Via the diverse pre-trained knowledge, they are also training-free and do not need any 3D training data. As compared in Table~\ref{supp_t9}, different from other methods based on 2D or 3D pre-training, our method is a pure non-parametric network without any learnable parameters or pre-trained knowledge.

\vspace{0.2cm}
\subsection{Point-Memory Bank vs. $k$-NN?} 
\vspace{0.2cm}
Based on the already extracted point cloud features, our point-memory bank and $k$-NN algorithm both leverage the inter-sample feature similarity for classification without training, but are different from the following two aspects.

\vspace{-0.2cm}
\paragraph{Soft Integration vs. Hard Assignment.}
As illustrated in Section \eqref{s3.3} of the main paper, our point-memory bank regards the similarities $S_{cos}$ between the test point cloud feature and the feature memory, $F_{mem}$, as weights, which are adopted for weighted summation of the one-hot label memory, $T_{mem}$. This can be viewed as a soft label integration. Instead, $k$-NN utilizes $S_{cos}$ to search the $k$ nearest neighbors from the training set, and directly outputs the category label with the maximum number of samples within the $k$ neighbors. Hence, $k$-NN conducts a hard label assignment, which is less adaptive than the soft integration. Additionally, our point-memory bank can be accomplished simply by two matrix multiplications and requires no sorting, which is more efficient for hardware implementation.

\paragraph{All Samples vs. $k$ Neighbors.}
Our point-memory bank integrates the entire label memory with different weights. This can take the semantics of all training samples into account for classification. In contrast, $k$-NN only involves the nearest $k$ neighbors to the test sample, which discards the sufficient category knowledge from other training samples. 

\begin{table}[t]
\centering
\caption{\textbf{Point-Memory Bank vs. $k$-NN.} `Top-$k$ PoM' denotes the point-memory bank with top-$k$ similarities, and `All' denotes 9,840 training samples. We utilize our non-parametric encoder to extract features and report the accuracy (\%) on ModelNet40~\cite{modelnet40}.}
\begin{adjustbox}{width=0.92\linewidth}
	\begin{tabular}{cccccccc}
	\toprule
	\makecell*[c]{$k$} &1 &10 &100 &500 &1000 &5000 &All\\
        \cmidrule(lr){1-1} \cmidrule(lr){2-8}
        \specialrule{0em}{1pt}{1pt}
		 Top-$k$ PoM &80.4 &81.1  &81.3  &81.4 &81.4 &81.7 &\textbf{81.8}\\
		 \specialrule{0em}{1pt}{1pt}
         $k$-NN &80.7 &79.5 &67.0  &45.7 &36.4 &8.5 &-\\ 
		 \specialrule{0em}{1pt}{1pt}
	\bottomrule
	\end{tabular}
\end{adjustbox}
\label{supp_t7}
\end{table}

\paragraph{Performance Comparison.}
In Table~\ref{supp_t7}, based on the point cloud features extracted by our non-parametric encoder, we implement the top-$k$ version of point-memory bank for comparison with $k$-NN, which only aggregates the label memory of the training samples with top-$k$ similarities. 
As the neighbor number $k$ increases, $k$-NN's performance is severely harmed due to its hard label assignment, while our point-memory bank attains the highest accuracy by utilizing all 9,840 samples for classification, indicating their different characters.

\vspace{0.2cm}
\subsection{Point-NN vs. PnP-3D?}
\vspace{0.2cm}
One previous work, PnP-3D~\cite{qiu2021pnp}, proposes local-global 3D processing modules that are plugged into other 3D models for performance improvement. Different from Point-NN's plug-and-play, PnP-3D introduces additional learnable parameters and requires to re-train the baseline networks from scratch, which is time-consuming. In contrast, our Point-NN is non-parametric and enhances the baseline directly during inference. In Table~\ref{supp_t10}, we compare Point-NN with PnP-3D respectively on PAConv~\cite{xu2021paconv} for shape classification and VoteNet~\cite{votenet} for 3D object detection. As shown, our method contributes to similar performance enhancement on the benchmarks, while brings no extra parameters or re-training.
In the table, we report the additional time for Point-NN to construct the point-memory bank before plug-and-play, which are 48 seconds and 9.3 minutes for the two tasks.

\section{Conclusion}
We revisit the non-learnable components in existing 3D models and propose Point-NN, a pure non-parametric network for 3D point cloud analysis. Free from any parameters or training, Point-NN achieves favorable accuracy on various 3D tasks. Starting from Point-NN, we propose its two promising applications: architectural frameworks for Point-PN and plug-and-play modules for performance improvement. Extensive experiments have demonstrated its effectiveness and significance. For future works, we will focus on exploring more advanced non-parametric models with wider application scenarios for 3D point cloud analysis.

\paragraph{Acknowledgement.}
This project is funded in part by National Key R\&D Program of China Project 2022ZD0161100, by the Centre for Perceptual and Interactive Intelligence (CPII) Ltd under the Innovation and Technology Commission (ITC)'s InnoHK, by General Research Fund of Hong Kong RGC Project 14204021, by the National Natural Science Foundation of China (Grant No. 62206272), and by Shanghai Committee of Science and Technology (Grant No. 21DZ1100100). 

\begin{table}[t!]
\centering
\small
\begin{adjustbox}{width=\linewidth}
	\begin{tabular}{c c c c c}
	\toprule
	Baseline &Method &Gain (\%) &Param. &Time\\
	\cmidrule(lr){1-1} \cmidrule(lr){2-2} \cmidrule(lr){3-3} \cmidrule(lr){4-4} \cmidrule(lr){5-5}
	\multirow{2}{*}{PAConv} &PnP-3D &+0.2 Acc.&+0.7 M &+14 h\\
        &Point-NN &+0.2 Acc.&+0 M &+48 s\\
        \midrule
        \multirow{2}{*}{VoteNet} &PnP-3D &+1.4 AP$_{25}$ &+0.3 M &+10 h\\
        &Point-NN &+1.2 AP$_{25}$ &+0 M &+9.3 min\\
	\bottomrule
	\end{tabular}
\end{adjustbox}
\caption{\textbf{Point-NN vs. PnP-3D~\cite{qiu2021pnp}.} We adopt two baseline models for comparison, PAConv~\cite{xu2021paconv} and VoteNet~\cite{votenet}, respectively on ModelNet40~\cite{modelnet40} and SUN RGB-D~\cite{sun_rgb} datasets.}
\label{supp_t10}
\end{table}

\clearpage
\vspace{0.5cm}

\appendix

\section{Related Work}
\label{s1}

\paragraph{3D Point Cloud Analysis.}
As the main data form in 3D, point clouds have stimulated a range of challenging tasks, including shape classification~\cite{qi2017pointnet,qi2017pointnet++,pointmlp,qian2022pointnext,liu2020closer, xie2021generative}, scene segmentation~\cite{lai2022stratified, dai20183dmv, zhao2019dar}, 3D object detection~\cite{3detr,votenet,shi2020pv, he2020structure,huang2022tig,zhang2022monodetr}, 3D vision-language learning~\cite{zhang2022pointclip,zhu2022pointclip,guo2022calip,wu2022eda}. Existing solutions as backbone networks can be categorized into projection-based and point-based methods.  To handle the irregularity and sparsity of point clouds, projection-based methods convert them into grid-like data, such as tangent
planes~\cite{tatarchenko2018tangent}, multi-view depth maps~\cite{hamdi2021mvtn,simpleview,zhang2022pointclip,su2015multi,zhu2022pointclip}, and 3D voxels~\cite{pvcnn,riegler2017octnet, meng2019vv,zhang2021dspoint}.  By doing this, the efficient 2D networks~\cite{resnet} and 3D convolutions~\cite{pvcnn} can be adopted for robust point cloud understanding. However, the projection process inevitably causes geometric information loss and quantization error. Point-based methods directly extract 3D patterns upon the unstructured input points to alleviate this loss of information. The seminal PointNet~\cite{qi2017pointnet} utilizes shared MLP layers to independently extract point features and aggregate the global representation via a max pooling. PointNet++~\cite{qi2017pointnet++} further constructs a multi-stage hierarchy to encode local spatial geometries progressively. Since then, the follow-up methods introduce advanced yet complicated local operators~\cite{xu2021paconv,pointmlp} and global transformers~\cite{guo2021pct,zhang2022point,zhang2022learning,guo2023joint,fu2022distillation,fu2022pos,chen2023pimae} for spatial geometry learning. In this paper, we follow the paradigm of more popular point-based methods, and propose a pure non-parametric network, Point-NN, with its two promising applications. For the first time, we verify the effectiveness of non-parametric components for 3D point cloud analysis.

\paragraph{Local Geometry Operators.}
Referring to the inductive bias of locality~\cite{resnet,krizhevsky2017imagenet}, most existing 3D models adopt delicate 3D operators to iteratively aggregate neighborhood features. Following PointNet++~\cite{qi2017pointnet++}, a series of methods utilize shared MLP layers with learnable relation modules for local pattern encoding, e.g., fully-linked webs~\cite{zhao2019pointweb}, structural relation network~\cite{duan2019structural}, and geometric affine module~\cite{pointmlp}. Some methods define irregular spatial kernels and introduce point-wise convolutions by relation mapping~\cite{rscnn}, Monte Carlo estimation~\cite{wu2019pointconv,hermosilla2018monte}, and dynamic kernel assembling~\cite{xu2021paconv}. Inspired by graph networks, DGCNN~\cite{dgcnn} and others~\cite{landrieu2018large,te2018rgcnn} regard points as vertices and interact local geometry through edges. Transformers~\cite{lai2022stratified,zhao2021point} are also introduced in 3D for attention-based feature communication. CurveNet~\cite{curvenet} proposes generating hypothetical curves for point grouping and feature aggregation. Unlike all previous methods with learnable operators, Point-NN adopts non-parametric trigonometric functions to reveal the spatial geometry within local regions, and Point-PN further appends simple linear layers on top with high performance-parameter trade-off.

\paragraph{Positional Encodings.}
Transformers~\cite{transformer} represent input signals as an orderless sequence and implicitly utilize positional encodings (PE) to inject positional information. Typically, trigonometric functions are adopted as the non-learnable PE~\cite{gehring2017convolutional} to encode both absolute and relative positions, each dimension corresponding to a sinusoid. For vision, PE can also be learnable during training~\cite{dosovitskiy2020image} or online predicted by neural networks~\cite{zhao2021point,chu2021conditional}. Another work~\cite{rahaman2019spectral} indicates that deep networks can learn better high-frequency variation given a higher dimensional input. Tancik \etal~\cite{tancik2020fourier} interpret it as Fourier transform to learn high-frequency functions in low dimensional problems. NeRF~\cite{mildenhall2021nerf} utilizes trigonometric PE to enhance the MLPs for better neural scene representations, but in a different formulation from the Transformer's. In contrast, we extend the non-learnable trigonometric PE of Transformer for specialized raw-point embedding and local geometry extraction, other than serving as an accessory in previous learnable networks. By doing this, the non-parametric encoder of Point-NN can effectively capture low-level spatial patterns complementary to the already trained 3D models.

\begin{figure*}[t!]
  \centering
  \vspace{0.3cm}
    \includegraphics[width=1\textwidth]{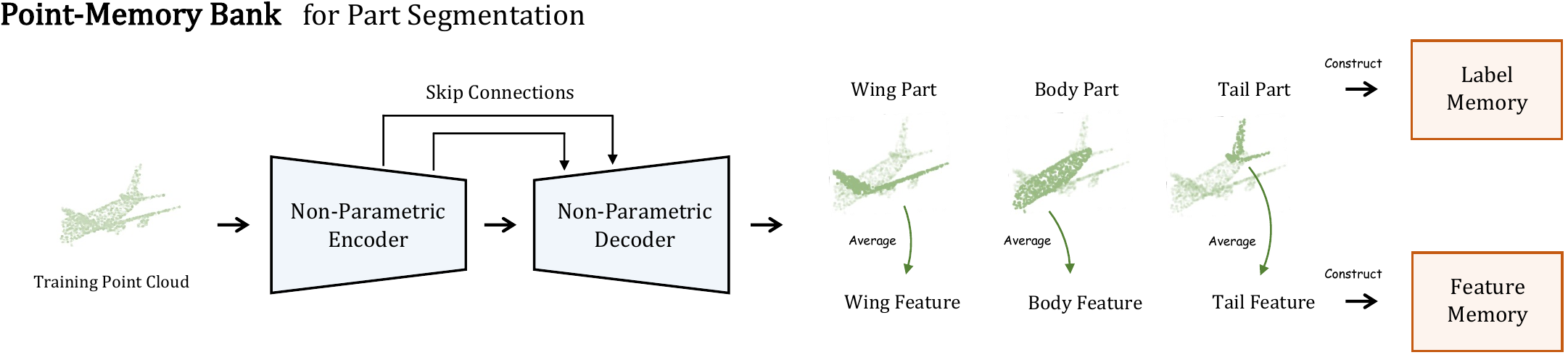}
   \caption{\textbf{Point-Memory Bank for Part Segmentation.} We first utilize the non-parametric encoder and decoder to extract the point-wise features of the point cloud in the training set. Then, we average the point features with the same part label to obtain the part-wise features, and construct them as the feature memory.}
    \label{supp_f1}
    \vspace{0.2cm}
\end{figure*}

\vspace{0.1cm}
\section{Implementation Details}
\vspace{0.1cm}
\label{s2}

\paragraph{Point-NN.}
The non-parametric encoder of Point-NN contains 4 stages. Each stage reduces the point number by half via FPS, and doubles the feature dimension during feature expansion.
For shape classification, the initial feature dimension $C_I$ is set to 72, and the final dimension $C_G$ of global representation is 1,152. The neighbor number $k$ of $k$-NN is 90 for all stages. We set the two hyperparameters $\alpha, \beta$ in $\operatorname{PosE}(\cdot)$ as 1000 and 100, respectively, referring to Equation \eqref{equa6} and \eqref{equa7} in the main paper. For part segmentation, we extend the non-parametric encoder into 5 stages for fully aggregating multi-scale 3D features. We set the initial feature dimension $C_I$ as 144, and the neighbor number $k$ as 128. We appended a non-parametric decoder with skip connections in each stage, which concatenate the propagated point features in the decoder with the corresponding ones from the encoder. As shown in Figure~\ref{supp_f1}, we construct the segmentation point-memory bank by storing the part-wise features and labels from the training set, which largely saves the GPU memory. During inference, each point-wise feature of the test point cloud conducts similarity matching with the part-wise feature memory for segmentation.

\begin{table}[t]
\small
\vspace{0.1cm}
\tabcaption{\textbf{Ablation Study of Non-Parametric Encoder.} We ablate the grouping method for local neighbors, feature expansion by concatenation, and pooling operation for feature aggregation. We report the classification accuracy (\%) on ModelNet40~\cite{modelnet40}.}
\begin{minipage}[t!]{0.32\linewidth}
\centering
\begin{adjustbox}{width=\linewidth}
	\begin{tabular}{cc}
	\toprule
	\makecell*[c]{Grouping\\Method} &Acc.\\
        \midrule
        Ball Query &78.5 \\
        \textbf{$k$-NN} &\textbf{81.8} \\
	\bottomrule
	\end{tabular}
\end{adjustbox}
\end{minipage}\,\,
\begin{minipage}[t!]{0.27\linewidth}
\centering
\begin{adjustbox}{width=\linewidth}
	\begin{tabular}{cc}
	\toprule
	\makecell*[c]{Feature\\Expand} &Acc.\\
        \midrule
        w/o &70.0 \\
        \textbf{w} &\textbf{81.8} \\
	\bottomrule
	\end{tabular}
\end{adjustbox}
\end{minipage}\,\,
\begin{minipage}[t!]{0.27\linewidth}
\centering
\begin{adjustbox}{width=\linewidth}
	\begin{tabular}{cc}
	\toprule
	\makecell*[c]{Pooling} &Acc.\\
        \midrule
        Max &80.4 \\
        Ave &77.1 \\
        \textbf{Both} &\textbf{81.8} \\
	\bottomrule
	\end{tabular}
\end{adjustbox}
\end{minipage}
\vspace{0.1cm}
\label{supp_t1}
\end{table}

\paragraph{Point-PN.}
For the parametric variant, we decrease $C_I$ to 36 and the neighbor number $k$ to 40 for lightweight parameters and efficient inference. We adopt the bottleneck architecture with a ratio 0.5 for the two cascaded linear layers after the \textit{Geometry Extraction} step. The initial parametric raw-point embedding consists of only one linear layer, and the final classifier contains three linear layers as existing methods~\cite{pointmlp,qian2022pointnext}. Specially, for the second 2-layer linear layers, i.e., the `2' of `1+2', at the first stage of Point-PN, we stack them twice for better extracting elementary 3D patterns at shallow layers. 
For shape classification, we train Point-PN for 300 epochs with a batch size 32 on a single RTX 3090 GPU. On ModelNet40~\cite{modelnet40}, we adopt SGD optimizer with a weight decay 0.0002, and cosine scheduler with an initial learning rate 0.1. On ScanObjectNN~\cite{scanobjectnn}, we adopt AdamW optimizer~\cite{kingma2014adam} with a weight decay 0.05, and cosine scheduler with an initial learning rate 0.002. We follow the data augmentation in PointMLP~\cite{pointmlp} and PointNeXt~\cite{qian2022pointnext} respectively for ModeNet40 and ScenObjectNN datasets. For part segmentation, we simply utilize the same learnbable decoder and training settings as CurveNet~\cite{curvenet} for a fair comparison.

\paragraph{Plug-and-play.}
For part segmentation and 3D object detection, concurrently running an extra Point-NN to enhance existing models would be expensive in both time and memory. Thus, referring to SN-Adapter~\cite{zhang2023nearest}, we directly adopt the encoders of already trained models to extract point cloud features, and only apply our point-memory bank on top for plug-and-play. In this way, we can also achieve performance improvement by leveraging the complementary knowledge between similarity matching and traditional learnable classification heads.

\section{Additional Ablation Study}
\label{s3}

\paragraph{Non-Parametric Encoder.}
In Table~\ref{supp_t1}, we further investigate other designs at every stage of Point-NN's non-parametric encoder. As shown, $k$-NN performs better than ball query~\cite{qi2017pointnet++} for grouping the neighbors of each center point since the ball query would fail to aggregate valid geometry in some sparse regions with only a few neighboring points.
Expanding the feature dimension by concatenating the center and neighboring points can improve the performance by +5.3\%. This is because each point obtains larger receptive fields as the network stage goes deeper and requires higher-dimensional vectors to encode more spatial semantics.  For the pooling operation after geometry extraction, we observe applying both max and average pooling achieves the highest accuracy, which can summarize the local patterns from two different aspects.

\begin{table}[t]
\centering
\vspace{0.1cm}
\caption{\textbf{Magnitude $\alpha$ in Trigonometric Functions.} We report the classification accuracy of Point-NN on ModelNet40~\cite{modelnet40}.\vspace{0.1cm}}
\begin{adjustbox}{width=0.96\linewidth}
	\begin{tabular}{lcccccc}
	\toprule
	\makecell*[c]{Magnitude $\alpha$} &\ 1\ &10 &50 &\textbf{100} &200 &500\\
        \cmidrule(lr){1-1} \cmidrule(lr){2-7}
        \specialrule{0em}{1pt}{1pt}
		 Acc. (\%) &\ 68.3\ &77.9  &81.1  &\textbf{81.8} &81.4 &81.0\ \\ 
		 \specialrule{0em}{1pt}{1pt}
	\bottomrule
	\end{tabular}
\end{adjustbox}
\label{supp_t2}
\end{table}
\begin{table}[t]
\centering
\caption{\textbf{Wavelength $\beta$ in Trigonometric Functions.} We report the classification accuracy of Point-NN on ModelNet40~\cite{modelnet40}.\vspace{0.2cm}}
\begin{adjustbox}{width=\linewidth}
	\begin{tabular}{lcccccc}
	\toprule
	\makecell*[c]{Wavelength $\beta$} &\ 10\ &100 &\textbf{500} &1000 &2000 &3000\ \\
        \cmidrule(lr){1-1} \cmidrule(lr){2-7}
        \specialrule{0em}{1pt}{1pt}
		 Acc. (\%) &\ 51.3\  &80.4 &\textbf{81.8} &74.5 &73.1 &72.9\ \\ 
		 \specialrule{0em}{1pt}{1pt}
	\bottomrule
	\end{tabular}
\end{adjustbox}
\label{supp_t3}
\end{table}

\paragraph{Hyperparameters in Trigonometric Functions.}
In Table~\ref{supp_t2} and \ref{supp_t3}, we show the influence of two hyperparameters in trigonometric functions of Point-NN. We fix one of them to be the best-performing value ($\alpha$ as 100, $\beta$ as 500), and vary the other one for ablation.
The combination of the magnitude $\alpha$ and wavelength $\beta$ control the frequency of the channel-wise sinusoid, and thus determine the raw point encoding for different classification accuracy.

\paragraph{Point-Memory Bank with Different Sizes.}
As default, we construct the feature memory by the entire training-set point clouds. In Table~\ref{supp_t4}, we report how Point-NN performs when partial training samples are utilized for the point-memory bank. As shown, Point-NN can attain 60.1\% classification accuracy with only 10\% of the training data, and further achieves 70.1\% with 40\% data, which is comparable to the performance of 100\% ratio but consumes less GPU memory. This indicates Point-NN is not sensitive to the memory bank size and can perform favorably with partial training-set data.

\begin{table}[t]
\centering
\caption{\textbf{Point-Memory Bank with Different Sizes.} We randomly sample different ratios of ModelNet40~\cite{modelnet40} to construct the point-memory bank and report the classification accuracy with GPU memory consumption.}
\vspace{0.2cm}
\begin{adjustbox}{width=\linewidth}
	\begin{tabular}{lccccccc}
	\toprule
	\makecell*[c]{Ratio (\%)} &1 &5 &10 &20 &40 &80 &100\\
        \cmidrule(lr){1-1} \cmidrule(lr){2-8}
        \specialrule{0em}{1pt}{1pt}
		 Acc. (\%)&39.5 &64.2  &70.3  &75.0 &77.9 &80.8 &\textbf{81.8}\\ 
		 \specialrule{0em}{1pt}{1pt}
         Mem. (G) &3.84 &3.87 &3.93  &4.05 &4.26 &4.82 &5.21\\ 
		 \specialrule{0em}{1pt}{1pt}
	\bottomrule
	\end{tabular}
\end{adjustbox}
\label{supp_t4}
\end{table}

\vspace{0.4cm}

{\small
\bibliographystyle{ieee_fullname}
\bibliography{egbib}

\begin{thebibliography}{10}\itemsep=-1pt

\bibitem{aldoma2012tutorial}
Aitor Aldoma, Zoltan-Csaba Marton, Federico Tombari, Walter Wohlkinger,
  Christian Potthast, Bernhard Zeisl, Radu~Bogdan Rusu, Suat Gedikli, and
  Markus Vincze.
\newblock Tutorial: Point cloud library: Three-dimensional object recognition
  and 6 dof pose estimation.
\newblock {\em IEEE Robotics \& Automation Magazine}, 19(3):80--91, 2012.

\bibitem{3dmfv}
Yizhak Ben-Shabat, Michael Lindenbaum, and Anath Fischer.
\newblock 3dmfv: Three-dimensional point cloud classification in real-time
  using convolutional neural networks.
\newblock {\em IEEE Robotics and Automation Letters}, 3(4):3145--3152, 2018.

\bibitem{chen2023pimae}
Anthony Chen, Kevin Zhang, Renrui Zhang, Zihan Wang, Yuheng Lu, Yandong Guo,
  and Shanghang Zhang.
\newblock Pimae: Point cloud and image interactive masked autoencoders for 3d
  object detection.
\newblock {\em CVPR 2023}, 2023.

\bibitem{chen2019deep}
Jingdao Chen, Zsolt Kira, and Yong~K Cho.
\newblock Deep learning approach to point cloud scene understanding for
  automated scan to 3d reconstruction.
\newblock {\em Journal of Computing in Civil Engineering}, 33(4):04019027,
  2019.

\bibitem{chen2017multi}
Xiaozhi Chen, Huimin Ma, Ji Wan, Bo Li, and Tian Xia.
\newblock Multi-view 3d object detection network for autonomous driving.
\newblock In {\em Proceedings of the IEEE conference on Computer Vision and
  Pattern Recognition}, pages 1907--1915, 2017.

\bibitem{chu2021conditional}
Xiangxiang Chu, Zhi Tian, Bo Zhang, Xinlong Wang, Xiaolin Wei, Huaxia Xia, and
  Chunhua Shen.
\newblock Conditional positional encodings for vision transformers.
\newblock {\em arXiv preprint arXiv:2102.10882}, 2021.

\bibitem{correll2016analysis}
Nikolaus Correll, Kostas~E Bekris, Dmitry Berenson, Oliver Brock, Albert Causo,
  Kris Hauser, Kei Okada, Alberto Rodriguez, Joseph~M Romano, and Peter~R
  Wurman.
\newblock Analysis and observations from the first amazon picking challenge.
\newblock {\em IEEE Transactions on Automation Science and Engineering},
  15(1):172--188, 2016.

\bibitem{cristianini2000introduction}
Nello Cristianini, John Shawe-Taylor, et~al.
\newblock {\em An introduction to support vector machines and other
  kernel-based learning methods}.
\newblock Cambridge university press, 2000.

\bibitem{ScanNetV2}
Angela Dai, Angel~X Chang, Manolis Savva, Maciej Halber, Thomas Funkhouser, and
  Matthias Nie{\ss}ner.
\newblock Scannet: Richly-annotated 3d reconstructions of indoor scenes.
\newblock In {\em Proceedings of the IEEE conference on computer vision and
  pattern recognition}, pages 5828--5839, 2017.

\bibitem{dai20183dmv}
Angela Dai and Matthias Nie{\ss}ner.
\newblock 3dmv: Joint 3d-multi-view prediction for 3d semantic scene
  segmentation.
\newblock In {\em Proceedings of the European Conference on Computer Vision
  (ECCV)}, pages 452--468, 2018.

\bibitem{votenet}
Zhipeng Ding, Xu Han, and Marc Niethammer.
\newblock Votenet: A deep learning label fusion method for multi-atlas
  segmentation.
\newblock In {\em International Conference on Medical Image Computing and
  Computer-Assisted Intervention}, pages 202--210. Springer, 2019.

\bibitem{dosovitskiy2020image}
Alexey Dosovitskiy, Lucas Beyer, Alexander Kolesnikov, Dirk Weissenborn,
  Xiaohua Zhai, Thomas Unterthiner, Mostafa Dehghani, Matthias Minderer, Georg
  Heigold, Sylvain Gelly, et~al.
\newblock An image is worth 16x16 words: Transformers for image recognition at
  scale.
\newblock {\em arXiv preprint arXiv:2010.11929}, 2020.

\bibitem{duan2019structural}
Yueqi Duan, Yu Zheng, Jiwen Lu, Jie Zhou, and Qi Tian.
\newblock Structural relational reasoning of point clouds.
\newblock In {\em Proceedings of the IEEE/CVF Conference on Computer Vision and
  Pattern Recognition}, pages 949--958, 2019.

\bibitem{friedman2001greedy}
Jerome~H Friedman.
\newblock Greedy function approximation: a gradient boosting machine.
\newblock {\em Annals of statistics}, pages 1189--1232, 2001.

\bibitem{fu2022pos}
Kexue Fu, Peng Gao, ShaoLei Liu, Renrui Zhang, Yu Qiao, and Manning Wang.
\newblock Pos-bert: Point cloud one-stage bert pre-training.
\newblock {\em arXiv preprint arXiv:2204.00989}, 2022.

\bibitem{fu2022distillation}
Kexue Fu, Peng Gao, Renrui Zhang, Hongsheng Li, Yu Qiao, and Manning Wang.
\newblock Distillation with contrast is all you need for self-supervised point
  cloud representation learning.
\newblock {\em arXiv preprint arXiv:2202.04241}, 2022.

\bibitem{gehring2017convolutional}
Jonas Gehring, Michael Auli, David Grangier, Denis Yarats, and Yann~N Dauphin.
\newblock Convolutional sequence to sequence learning.
\newblock In {\em International conference on machine learning}, pages
  1243--1252. PMLR, 2017.

\bibitem{simpleview}
Ankit Goyal, Hei Law, Bowei Liu, Alejandro Newell, and Jia Deng.
\newblock Revisiting point cloud shape classification with a simple and
  effective baseline.
\newblock {\em arXiv preprint arXiv:2106.05304}, 2021.

\bibitem{guo2021pct}
Meng-Hao Guo, Jun-Xiong Cai, Zheng-Ning Liu, Tai-Jiang Mu, Ralph~R Martin, and
  Shi-Min Hu.
\newblock Pct: Point cloud transformer.
\newblock {\em Computational Visual Media}, 7(2):187--199, 2021.

\bibitem{guo2023joint}
Ziyu Guo, Xianzhi Li, and Pheng~Ann Heng.
\newblock Joint-mae: 2d-3d joint masked autoencoders for 3d point cloud
  pre-training.
\newblock {\em arXiv preprint arXiv:2302.14007}, 2023.

\bibitem{guo2022calip}
Ziyu Guo, Renrui Zhang, Longtian Qiu, Xianzheng Ma, Xupeng Miao, Xuming He, and
  Bin Cui.
\newblock Calip: Zero-shot enhancement of clip with parameter-free attention.
\newblock {\em arXiv preprint arXiv:2209.14169}, 2022.

\bibitem{hamdi2021mvtn}
Abdullah Hamdi, Silvio Giancola, and Bernard Ghanem.
\newblock Mvtn: Multi-view transformation network for 3d shape recognition.
\newblock In {\em Proceedings of the IEEE/CVF International Conference on
  Computer Vision}, pages 1--11, 2021.

\bibitem{he2020structure}
Chenhang He, Hui Zeng, Jianqiang Huang, Xian-Sheng Hua, and Lei Zhang.
\newblock Structure aware single-stage 3d object detection from point cloud.
\newblock In {\em Proceedings of the IEEE/CVF conference on computer vision and
  pattern recognition}, pages 11873--11882, 2020.

\bibitem{resnet}
Kaiming He, Xiangyu Zhang, Shaoqing Ren, and Jian Sun.
\newblock Deep residual learning for image recognition.
\newblock In {\em Proceedings of the IEEE conference on computer vision and
  pattern recognition}, pages 770--778, 2016.

\bibitem{hermosilla2018monte}
Pedro Hermosilla, Tobias Ritschel, Pere-Pau V{\'a}zquez, {\`A}lvar Vinacua, and
  Timo Ropinski.
\newblock Monte carlo convolution for learning on non-uniformly sampled point
  clouds.
\newblock {\em ACM Transactions on Graphics (TOG)}, 37(6):1--12, 2018.

\bibitem{huang2022tig}
Peixiang Huang, Li Liu, Renrui Zhang, Song Zhang, Xinli Xu, Baichao Wang, and
  Guoyi Liu.
\newblock Tig-bev: Multi-view bev 3d object detection via target inner-geometry
  learning.
\newblock {\em arXiv preprint arXiv:2212.13979}, 2022.

\bibitem{huang2022clip2point}
Tianyu Huang, Bowen Dong, Yunhan Yang, Xiaoshui Huang, Rynson~WH Lau, Wanli
  Ouyang, and Wangmeng Zuo.
\newblock Clip2point: Transfer clip to point cloud classification with
  image-depth pre-training.
\newblock {\em arXiv preprint arXiv:2210.01055}, 2022.

\bibitem{kidono2011pedestrian}
Kiyosumi Kidono, Takeo Miyasaka, Akihiro Watanabe, Takashi Naito, and Jun
  Miura.
\newblock Pedestrian recognition using high-definition lidar.
\newblock In {\em 2011 IEEE Intelligent Vehicles Symposium (IV)}, pages
  405--410. IEEE, 2011.

\bibitem{kingma2014adam}
Diederik~P Kingma and Jimmy Ba.
\newblock Adam: A method for stochastic optimization.
\newblock {\em arXiv preprint arXiv:1412.6980}, 2014.

\bibitem{krizhevsky2017imagenet}
Alex Krizhevsky, Ilya Sutskever, and Geoffrey~E Hinton.
\newblock Imagenet classification with deep convolutional neural networks.
\newblock {\em Communications of the ACM}, 60(6):84--90, 2017.

\bibitem{lai2022stratified}
Xin Lai, Jianhui Liu, Li Jiang, Liwei Wang, Hengshuang Zhao, Shu Liu, Xiaojuan
  Qi, and Jiaya Jia.
\newblock Stratified transformer for 3d point cloud segmentation.
\newblock In {\em Proceedings of the IEEE/CVF Conference on Computer Vision and
  Pattern Recognition}, pages 8500--8509, 2022.

\bibitem{landrieu2018large}
Loic Landrieu and Martin Simonovsky.
\newblock Large-scale point cloud semantic segmentation with superpoint graphs.
\newblock In {\em Proceedings of the IEEE conference on computer vision and
  pattern recognition}, pages 4558--4567, 2018.

\bibitem{li2018pointcnn}
Yangyan Li, Rui Bu, Mingchao Sun, Wei Wu, Xinhan Di, and Baoquan Chen.
\newblock Pointcnn: Convolution on x-transformed points.
\newblock {\em Advances in neural information processing systems}, 31:820--830,
  2018.

\bibitem{densepoint}
Yongcheng Liu, Bin Fan, Gaofeng Meng, Jiwen Lu, Shiming Xiang, and Chunhong
  Pan.
\newblock Densepoint: Learning densely contextual representation for efficient
  point cloud processing.
\newblock In {\em Proceedings of the IEEE/CVF International Conference on
  Computer Vision}, pages 5239--5248, 2019.

\bibitem{rscnn}
Yongcheng Liu, Bin Fan, Shiming Xiang, and Chunhong Pan.
\newblock Relation-shape convolutional neural network for point cloud analysis.
\newblock In {\em Proceedings of the IEEE/CVF Conference on Computer Vision and
  Pattern Recognition}, pages 8895--8904, 2019.

\bibitem{liu2020closer}
Ze Liu, Han Hu, Yue Cao, Zheng Zhang, and Xin Tong.
\newblock A closer look at local aggregation operators in point cloud analysis.
\newblock In {\em Computer Vision--ECCV 2020: 16th European Conference,
  Glasgow, UK, August 23--28, 2020, Proceedings, Part XXIII 16}, pages
  326--342. Springer, 2020.

\bibitem{pvcnn}
Zhijian Liu, Haotian Tang, Yujun Lin, and Song Han.
\newblock Point-voxel cnn for efficient 3d deep learning.
\newblock {\em arXiv preprint arXiv:1907.03739}, 2019.

\bibitem{pointmlp}
Xu Ma, Can Qin, Haoxuan You, Haoxi Ran, and Yun Fu.
\newblock Rethinking network design and local geometry in point cloud: A simple
  residual mlp framework.
\newblock {\em arXiv preprint arXiv:2202.07123}, 2022.

\bibitem{meng2019vv}
Hsien-Yu Meng, Lin Gao, Yu-Kun Lai, and Dinesh Manocha.
\newblock Vv-net: Voxel vae net with group convolutions for point cloud
  segmentation.
\newblock In {\em Proceedings of the IEEE/CVF international conference on
  computer vision}, pages 8500--8508, 2019.

\bibitem{mildenhall2021nerf}
Ben Mildenhall, Pratul~P Srinivasan, Matthew Tancik, Jonathan~T Barron, Ravi
  Ramamoorthi, and Ren Ng.
\newblock Nerf: Representing scenes as neural radiance fields for view
  synthesis.
\newblock {\em Communications of the ACM}, 65(1):99--106, 2021.

\bibitem{3detr}
Ishan Misra, Rohit Girdhar, and Armand Joulin.
\newblock An end-to-end transformer model for 3d object detection.
\newblock In {\em Proceedings of the IEEE/CVF International Conference on
  Computer Vision (ICCV)}, pages 2906--2917, October 2021.

\bibitem{mousavian20196}
Arsalan Mousavian, Clemens Eppner, and Dieter Fox.
\newblock 6-dof graspnet: Variational grasp generation for object manipulation.
\newblock In {\em Proceedings of the IEEE/CVF International Conference on
  Computer Vision}, pages 2901--2910, 2019.

\bibitem{navarro2010pedestrian}
Luis~E Navarro-Serment, Christoph Mertz, and Martial Hebert.
\newblock Pedestrian detection and tracking using three-dimensional ladar data.
\newblock {\em The International Journal of Robotics Research},
  29(12):1516--1528, 2010.

\bibitem{qi2017pointnet}
Charles~R Qi, Hao Su, Kaichun Mo, and Leonidas~J Guibas.
\newblock Pointnet: Deep learning on point sets for 3d classification and
  segmentation.
\newblock In {\em Proceedings of the IEEE conference on computer vision and
  pattern recognition}, pages 652--660, 2017.

\bibitem{qi2017pointnet++}
Charles~R Qi, Li Yi, Hao Su, and Leonidas~J Guibas.
\newblock Pointnet++: Deep hierarchical feature learning on point sets in a
  metric space.
\newblock {\em arXiv preprint arXiv:1706.02413}, 2017.

\bibitem{qian2022pointnext}
Guocheng Qian, Yuchen Li, Houwen Peng, Jinjie Mai, Hasan Abed Al~Kader Hammoud,
  Mohamed Elhoseiny, and Bernard Ghanem.
\newblock Pointnext: Revisiting pointnet++ with improved training and scaling
  strategies.
\newblock {\em arXiv preprint arXiv:2206.04670}, 2022.

\bibitem{drnet}
Shi Qiu, Saeed Anwar, and Nick Barnes.
\newblock Dense-resolution network for point cloud classification and
  segmentation.
\newblock In {\em Proceedings of the IEEE/CVF Winter Conference on Applications
  of Computer Vision}, pages 3813--3822, 2021.

\bibitem{gbnet}
Shi Qiu, Saeed Anwar, and Nick Barnes.
\newblock Geometric back-projection network for point cloud classification.
\newblock {\em IEEE Transactions on Multimedia}, 2021.

\bibitem{qiu2021pnp}
Shi Qiu, Saeed Anwar, and Nick Barnes.
\newblock Pnp-3d: A plug-and-play for 3d point clouds.
\newblock {\em IEEE Transactions on Pattern Analysis and Machine Intelligence},
  45(1):1312--1319, 2021.

\bibitem{radford2021learning}
Alec Radford, Jong~Wook Kim, Chris Hallacy, Aditya Ramesh, Gabriel Goh,
  Sandhini Agarwal, Girish Sastry, Amanda Askell, Pamela Mishkin, Jack Clark,
  et~al.
\newblock Learning transferable visual models from natural language
  supervision.
\newblock In {\em International conference on machine learning}, pages
  8748--8763. PMLR, 2021.

\bibitem{rahaman2019spectral}
Nasim Rahaman, Aristide Baratin, Devansh Arpit, Felix Draxler, Min Lin, Fred
  Hamprecht, Yoshua Bengio, and Aaron Courville.
\newblock On the spectral bias of neural networks.
\newblock In {\em International Conference on Machine Learning}, pages
  5301--5310. PMLR, 2019.

\bibitem{riegler2017octnet}
Gernot Riegler, Ali Osman~Ulusoy, and Andreas Geiger.
\newblock Octnet: Learning deep 3d representations at high resolutions.
\newblock In {\em Proceedings of the IEEE conference on computer vision and
  pattern recognition}, pages 3577--3586, 2017.

\bibitem{rusu2009close}
Radu~Bogdan Rusu, Nico Blodow, Zoltan~Csaba Marton, and Michael Beetz.
\newblock Close-range scene segmentation and reconstruction of 3d point cloud
  maps for mobile manipulation in domestic environments.
\newblock In {\em 2009 IEEE/RSJ International Conference on Intelligent Robots
  and Systems}, pages 1--6. IEEE, 2009.

\bibitem{safavian1991survey}
S~Rasoul Safavian and David Landgrebe.
\newblock A survey of decision tree classifier methodology.
\newblock {\em IEEE transactions on systems, man, and cybernetics},
  21(3):660--674, 1991.

\bibitem{sharma2020self}
Charu Sharma and Manohar Kaul.
\newblock Self-supervised few-shot learning on point clouds.
\newblock {\em arXiv preprint arXiv:2009.14168}, 2020.

\bibitem{shi2020pv}
Shaoshuai Shi, Chaoxu Guo, Li Jiang, Zhe Wang, Jianping Shi, Xiaogang Wang, and
  Hongsheng Li.
\newblock Pv-rcnn: Point-voxel feature set abstraction for 3d object detection.
\newblock In {\em Proceedings of the IEEE/CVF Conference on Computer Vision and
  Pattern Recognition}, pages 10529--10538, 2020.

\bibitem{sun_rgb}
Shuran Song, Samuel~P Lichtenberg, and Jianxiong Xiao.
\newblock Sun rgb-d: A rgb-d scene understanding benchmark suite.
\newblock In {\em Proceedings of the IEEE conference on computer vision and
  pattern recognition}, pages 567--576, 2015.

\bibitem{su2015multi}
Hang Su, Subhransu Maji, Evangelos Kalogerakis, and Erik Learned-Miller.
\newblock Multi-view convolutional neural networks for 3d shape recognition.
\newblock In {\em Proceedings of the IEEE international conference on computer
  vision}, pages 945--953, 2015.

\bibitem{tancik2020fourier}
Matthew Tancik, Pratul Srinivasan, Ben Mildenhall, Sara Fridovich-Keil, Nithin
  Raghavan, Utkarsh Singhal, Ravi Ramamoorthi, Jonathan Barron, and Ren Ng.
\newblock Fourier features let networks learn high frequency functions in low
  dimensional domains.
\newblock {\em Advances in Neural Information Processing Systems},
  33:7537--7547, 2020.

\bibitem{tatarchenko2018tangent}
Maxim Tatarchenko, Jaesik Park, Vladlen Koltun, and Qian-Yi Zhou.
\newblock Tangent convolutions for dense prediction in 3d.
\newblock In {\em Proceedings of the IEEE Conference on Computer Vision and
  Pattern Recognition}, pages 3887--3896, 2018.

\bibitem{te2018rgcnn}
Gusi Te, Wei Hu, Amin Zheng, and Zongming Guo.
\newblock Rgcnn: Regularized graph cnn for point cloud segmentation.
\newblock In {\em Proceedings of the 26th ACM international conference on
  Multimedia}, pages 746--754, 2018.

\bibitem{thomas2019kpconv}
Hugues Thomas, Charles~R Qi, Jean-Emmanuel Deschaud, Beatriz Marcotegui,
  Fran{\c{c}}ois Goulette, and Leonidas~J Guibas.
\newblock Kpconv: Flexible and deformable convolution for point clouds.
\newblock In {\em Proceedings of the IEEE/CVF International Conference on
  Computer Vision}, pages 6411--6420, 2019.

\bibitem{scanobjectnn}
Mikaela~Angelina Uy, Quang-Hieu Pham, Binh-Son Hua, Thanh Nguyen, and Sai-Kit
  Yeung.
\newblock Revisiting point cloud classification: A new benchmark dataset and
  classification model on real-world data.
\newblock In {\em Proceedings of the IEEE/CVF International Conference on
  Computer Vision}, pages 1588--1597, 2019.

\bibitem{transformer}
Ashish Vaswani, Noam Shazeer, Niki Parmar, Jakob Uszkoreit, Llion Jones,
  Aidan~N Gomez, {\L}ukasz Kaiser, and Illia Polosukhin.
\newblock Attention is all you need.
\newblock In {\em Advances in neural information processing systems}, pages
  5998--6008, 2017.

\bibitem{verdoja2017fast}
Francesco Verdoja, Diego Thomas, and Akihiro Sugimoto.
\newblock Fast 3d point cloud segmentation using supervoxels with geometry and
  color for 3d scene understanding.
\newblock In {\em 2017 IEEE International Conference on Multimedia and Expo
  (ICME)}, pages 1285--1290. IEEE, 2017.

\bibitem{dgcnn}
Yue Wang, Yongbin Sun, Ziwei Liu, Sanjay~E Sarma, Michael~M Bronstein, and
  Justin~M Solomon.
\newblock Dynamic graph cnn for learning on point clouds.
\newblock {\em Acm Transactions On Graphics (tog)}, 38(5):1--12, 2019.

\bibitem{wu2016learning}
Jiajun Wu, Chengkai Zhang, Tianfan Xue, William~T Freeman, and Joshua~B
  Tenenbaum.
\newblock Learning a probabilistic latent space of object shapes via 3d
  generative-adversarial modeling.
\newblock In {\em Proceedings of the 30th International Conference on Neural
  Information Processing Systems}, pages 82--90, 2016.

\bibitem{wu2019pointconv}
Wenxuan Wu, Zhongang Qi, and Li Fuxin.
\newblock Pointconv: Deep convolutional networks on 3d point clouds.
\newblock In {\em Proceedings of the IEEE/CVF Conference on Computer Vision and
  Pattern Recognition}, pages 9621--9630, 2019.

\bibitem{wu2022eda}
Yanmin Wu, Xinhua Cheng, Renrui Zhang, Zesen Cheng, and Jian Zhang.
\newblock Eda: Explicit text-decoupling and dense alignment for 3d visual and
  language learning.
\newblock {\em arXiv preprint arXiv:2209.14941}, 2022.

\bibitem{modelnet40}
Zhirong Wu, Shuran Song, Aditya Khosla, Fisher Yu, Linguang Zhang, Xiaoou Tang,
  and Jianxiong Xiao.
\newblock 3d shapenets: A deep representation for volumetric shapes.
\newblock In {\em Proceedings of the IEEE conference on computer vision and
  pattern recognition}, pages 1912--1920, 2015.

\bibitem{curvenet}
Tiange Xiang, Chaoyi Zhang, Yang Song, Jianhui Yu, and Weidong Cai.
\newblock Walk in the cloud: Learning curves for point clouds shape analysis.
\newblock {\em arXiv preprint arXiv:2105.01288}, 2021.

\bibitem{xie2021generative}
Jianwen Xie, Yifei Xu, Zilong Zheng, Song-Chun Zhu, and Ying~Nian Wu.
\newblock Generative pointnet: Deep energy-based learning on unordered point
  sets for 3d generation, reconstruction and classification.
\newblock In {\em Proceedings of the IEEE/CVF Conference on Computer Vision and
  Pattern Recognition}, pages 14976--14985, 2021.

\bibitem{xu2021paconv}
Mutian Xu, Runyu Ding, Hengshuang Zhao, and Xiaojuan Qi.
\newblock Paconv: Position adaptive convolution with dynamic kernel assembling
  on point clouds.
\newblock In {\em Proceedings of the IEEE/CVF Conference on Computer Vision and
  Pattern Recognition}, pages 3173--3182, 2021.

\bibitem{xu2021learning}
Mutian Xu, Junhao Zhang, Zhipeng Zhou, Mingye Xu, Xiaojuan Qi, and Yu Qiao.
\newblock Learning geometry-disentangled representation for complementary
  understanding of 3d object point cloud.
\newblock In {\em Proceedings of the AAAI Conference on Artificial
  Intelligence}, volume~35, pages 3056--3064, 2021.

\bibitem{xu2018spidercnn}
Yifan Xu, Tianqi Fan, Mingye Xu, Long Zeng, and Yu Qiao.
\newblock Spidercnn: Deep learning on point sets with parameterized
  convolutional filters.
\newblock In {\em Proceedings of the European Conference on Computer Vision
  (ECCV)}, pages 87--102, 2018.

\bibitem{xue2022ulip}
Le Xue, Mingfei Gao, Chen Xing, Roberto Mart{\'\i}n-Mart{\'\i}n, Jiajun Wu,
  Caiming Xiong, Ran Xu, Juan~Carlos Niebles, and Silvio Savarese.
\newblock Ulip: Learning unified representation of language, image and point
  cloud for 3d understanding.
\newblock {\em arXiv preprint arXiv:2212.05171}, 2022.

\bibitem{yang2018foldingnet}
Yaoqing Yang, Chen Feng, Yiru Shen, and Dong Tian.
\newblock Foldingnet: Point cloud auto-encoder via deep grid deformation.
\newblock In {\em Proceedings of the IEEE conference on computer vision and
  pattern recognition}, pages 206--215, 2018.

\bibitem{shapenetpart}
Li Yi, Vladimir~G Kim, Duygu Ceylan, I-Chao Shen, Mengyan Yan, Hao Su, Cewu Lu,
  Qixing Huang, Alla Sheffer, and Leonidas Guibas.
\newblock A scalable active framework for region annotation in 3d shape
  collections.
\newblock {\em ACM Transactions on Graphics (ToG)}, 35(6):1--12, 2016.

\bibitem{zhang2021tip}
Renrui Zhang, Rongyao Fang, Peng Gao, Wei Zhang, Kunchang Li, Jifeng Dai, Yu
  Qiao, and Hongsheng Li.
\newblock Tip-adapter: Training-free clip-adapter for better vision-language
  modeling.
\newblock {\em arXiv preprint arXiv:2111.03930}, 2021.

\bibitem{zhang2022point}
Renrui Zhang, Ziyu Guo, Peng Gao, Rongyao Fang, Bin Zhao, Dong Wang, Yu Qiao,
  and Hongsheng Li.
\newblock Point-m2ae: Multi-scale masked autoencoders for hierarchical point
  cloud pre-training.
\newblock {\em arXiv preprint arXiv:2205.14401}, 2022.

\bibitem{zhang2022pointclip}
Renrui Zhang, Ziyu Guo, Wei Zhang, Kunchang Li, Xupeng Miao, Bin Cui, Yu Qiao,
  Peng Gao, and Hongsheng Li.
\newblock Pointclip: Point cloud understanding by clip.
\newblock In {\em Proceedings of the IEEE/CVF Conference on Computer Vision and
  Pattern Recognition}, pages 8552--8562, 2022.

\bibitem{zhang2022monodetr}
Renrui Zhang, Han Qiu, Tai Wang, Xuanzhuo Xu, Ziyu Guo, Yu Qiao, Peng Gao, and
  Hongsheng Li.
\newblock Monodetr: Depth-aware transformer for monocular 3d object detection.
\newblock {\em arXiv preprint arXiv:2203.13310}, 2022.

\bibitem{zhang2023nearest}
Renrui Zhang, Liuhui Wang, Ziyu Guo, and Jianbo Shi.
\newblock Nearest neighbors meet deep neural networks for point cloud analysis.
\newblock In {\em Proceedings of the IEEE/CVF Winter Conference on Applications
  of Computer Vision}, pages 1246--1255, 2023.

\bibitem{zhang2022learning}
Renrui Zhang, Liuhui Wang, Yu Qiao, Peng Gao, and Hongsheng Li.
\newblock Learning 3d representations from 2d pre-trained models via
  image-to-point masked autoencoders.
\newblock {\em arXiv preprint arXiv:2212.06785}, 2022.

\bibitem{zhang2021dspoint}
Renrui Zhang, Ziyao Zeng, Ziyu Guo, Xinben Gao, Kexue Fu, and Jianbo Shi.
\newblock Dspoint: Dual-scale point cloud recognition with high-frequency
  fusion.
\newblock {\em arXiv preprint arXiv:2111.10332}, 2021.

\bibitem{zhao2019pointweb}
Hengshuang Zhao, Li Jiang, Chi-Wing Fu, and Jiaya Jia.
\newblock Pointweb: Enhancing local neighborhood features for point cloud
  processing.
\newblock In {\em Proceedings of the IEEE/CVF conference on computer vision and
  pattern recognition}, pages 5565--5573, 2019.

\bibitem{zhao2021point}
Hengshuang Zhao, Li Jiang, Jiaya Jia, Philip~HS Torr, and Vladlen Koltun.
\newblock Point transformer.
\newblock In {\em Proceedings of the IEEE/CVF International Conference on
  Computer Vision}, pages 16259--16268, 2021.

\bibitem{zhao2019dar}
Zongyue Zhao, Min Liu, and Karthik Ramani.
\newblock Dar-net: Dynamic aggregation network for semantic scene segmentation.
\newblock {\em arXiv preprint arXiv:1907.12022}, 2019.

\bibitem{zheng2013beyond}
Bo Zheng, Yibiao Zhao, Joey~C Yu, Katsushi Ikeuchi, and Song-Chun Zhu.
\newblock Beyond point clouds: Scene understanding by reasoning geometry and
  physics.
\newblock In {\em Proceedings of the IEEE Conference on Computer Vision and
  Pattern Recognition}, pages 3127--3134, 2013.

\bibitem{zhu2022pointclip}
Xiangyang Zhu, Renrui Zhang, Bowei He, Ziyao Zeng, Shanghang Zhang, and Peng
  Gao.
\newblock Pointclip v2: Adapting clip for powerful 3d open-world learning.
\newblock {\em arXiv preprint arXiv:2211.11682}, 2022.

\end{thebibliography}
}

\end{document}


\title{Supplementary Material}

\author{First Author\\
Institution1\\
Institution1 address\\
{\tt\small firstauthor@i1.org}
\and
Second Author\\
Institution2\\
First line of institution2 address\\
{\tt\small secondauthor@i2.org}
}
\maketitle


\appendix

\section{Related Work}
\label{s1}

\paragraph{3D Point Cloud Analysis.}
As the main data form in 3D, point clouds have stimulated a range of challenging tasks, including shape classification~\cite{qi2017pointnet,qi2017pointnet++,pointmlp,zhang2022pointclip}, scene segmentation~\cite{lai2022stratified, dai20183dmv, zhao2019dar}, and 3D object detection~\cite{3detr,votenet,shi2020pv, he2020structure}. Existing solutions can be categorized into two groups: projection-based and point-based methods.  To handle the irregularity and sparsity of point clouds, projection-based methods convert them into grid-like data, such as tangent
planes~\cite{tatarchenko2018tangent}, multi-view depth maps~\cite{hamdi2021mvtn,simpleview,zhang2022pointclip,su2015multi}, and 3D voxels~\cite{pvcnn,riegler2017octnet, meng2019vv}.  By doing this, the efficient 2D networks~\cite{resnet} and 3D convolutions~\cite{pvcnn} can be adopted for robust point cloud understanding. However, the projection process inevitably causes geometric information loss and quantization error. Point-based methods directly extract 3D patterns upon the unstructured input points to alleviate this loss of information. The seminal PointNet~\cite{qi2017pointnet} utilizes shared MLP layers to independently extract point features and aggregate the global representation via a max pooling. PointNet++~\cite{qi2017pointnet++} further constructs a multi-stage hierarchy to encode local spatial geometries progressively. Since then, the follow-up methods introduce advanced yet complicated local operators~\cite{xu2021paconv,pointmlp} and network architectures~\cite{guo2021pct,dgcnn} for spatial geometry learning. In this paper, we follow the paradigm of more popular point-based methods, and propose a pure non-parametric network, Point-NN, with its two promising applications. For the first time, we verify the effectiveness of non-parametric components for 3D point cloud analysis.

\paragraph{Local Geometry Operators.}
Referring to the inductive bias of locality~\cite{resnet,krizhevsky2017imagenet}, most existing 3D models adopt delicate 3D operators to iteratively aggregate neighborhood features. Following PointNet++~\cite{qi2017pointnet++}, a series of methods utilize shared MLP layers with learnable relation modules for local pattern encoding, e.g., fully-linked webs~\cite{zhao2019pointweb}, structural relation network~\cite{duan2019structural}, and geometric affine module~\cite{pointmlp}. Some methods define irregular spatial kernels and introduce point-wise convolutions by relation mapping~\cite{rscnn}, Monte Carlo estimation~\cite{wu2019pointconv,hermosilla2018monte}, and dynamic kernel assembling~\cite{xu2021paconv}. Inspired by graph networks, DGCNN~\cite{dgcnn} and others~\cite{landrieu2018large,te2018rgcnn} regard points as vertices and interact local geometry through edges. Transformers~\cite{lai2022stratified,zhao2021point} are also introduced in 3D for attention-based feature communication. CurveNet~\cite{curvenet} proposes generating hypothetical curves for point grouping and feature aggregation. Unlike all previous methods with learnable parameters, Point-NN adopts non-parametric trigonometric functions to reveal the spatial geometry within local regions. Point-PN further appends simple linear layers on top and achieves a high performance-parameter trade-off without complicated operations.

\paragraph{Positional Encodings.}
Transformers~\cite{transformer} represent input signals as an orderless sequence and implicitly utilize positional encodings (PE) to inject positional information. Typically, trigonometric functions are adopted as the non-learnable PE~\cite{gehring2017convolutional} to encode both absolute and relative positions, each dimension corresponding to a sinusoid. For vision, PE can also be learnable during training~\cite{dosovitskiy2020image} or online predicted by neural networks~\cite{zhao2021point,chu2021conditional}. Another work~\cite{rahaman2019spectral} indicates that deep networks can learn better high-frequency variation given a higher dimensional input. NerF~\cite{mildenhall2021nerf} utilizes trigonometric PE to enhance the MLPs for better neural scene representations, but in a different formulation from the Transformer's. Tancik \etal~\cite{tancik2020fourier} further interpret it as Fourier transform to learn high-frequency functions in low dimensional problems. In contrast, we extend the non-learnable trigonometric PE of Transformer for specialized raw-point embedding and local geometry extraction, other than serving as an accessory in previous learnable networks. By doing this, the non-parametric encoder of Point-NN can effectively capture low-level spatial patterns complementary to the already trained 3D models.

\section{Implementation Details}
\label{s2}

\paragraph{Point-NN.}
The non-parametric encoder of Point-NN contains 4 stages. Each stage reduces the point number by half via FPS, and doubles the feature dimension during feature expansion.
The initial feature dimension $C_I$ is set to 72, and the final dimension $C_G$ of global representation is 1,152. The neighbor number $k$ of $k$-NN is 84 for all stages. We set the two hyperparameters $\alpha, \beta$ in $\operatorname{PosE}(\cdot)$ as 500 and 100, respectively, referring to Equation (6) and (7) in the main paper. For part segmentation, the appended non-parametric decoder also conducts feature expansion in each stage by skip connections referring to PointNet++~\cite{qi2017pointnet++}, which concatenate the propagated point features in the decoder with the corresponding ones from the encoder. As shown in Figure~\ref{supp_f1}, we construct the segmentation point-memory bank by storing the part-wise features and labels from the training set, which largely saves the GPU memory. During inference, each point-wise feature of the test point cloud conducts similarity matching with the part-wise feature memory for segmentation.

\begin{figure*}[t!]
  \centering
    \includegraphics[width=1\textwidth]{cvpr2023-author_kit-v1_1-1/latex/figs/sup-fig1.pdf}
   \caption{\textbf{Point-Memory Bank for Part Segmentation.} We first utilize the non-parametric encoder and decoder to extract the point-wise features of the point cloud in the training set. Then, we average the point features with the same part label to obtain the part-wise features, and construct them as the feature memory.}
    \label{supp_f1}
    \vspace{0.1cm}
\end{figure*}

\begin{table}[t]
\small
\tabcaption{\textbf{Ablation Study of Non-Parametric Encoder.} We ablate different designs at every stage of the encoder: the grouping method for local neighbors, feature expansion by concatenation, and pooling operation for feature aggregation. We report the classification accuracy (\%) on ModelNet40~\cite{modelnet40}.}
\begin{minipage}[t!]{0.32\linewidth}
\centering
\begin{adjustbox}{width=\linewidth}
	\begin{tabular}{cc}
	\toprule
	\makecell*[c]{Grouping\\Method} &Acc.\\
        \midrule
        Ball Query &67.1 \\
        \textbf{$k$-NN} &\textbf{76.9} \\
	\bottomrule
	\end{tabular}
\end{adjustbox}
\end{minipage}\,\,
\begin{minipage}[t!]{0.27\linewidth}
\centering
\begin{adjustbox}{width=\linewidth}
	\begin{tabular}{cc}
	\toprule
	\makecell*[c]{Feature\\Expand} &Acc.\\
        \midrule
        w/o &71.6 \\
        \textbf{w} &\textbf{76.9} \\
	\bottomrule
	\end{tabular}
\end{adjustbox}
\end{minipage}\,\,
\begin{minipage}[t!]{0.27\linewidth}
\centering
\begin{adjustbox}{width=\linewidth}
	\begin{tabular}{cc}
	\toprule
	\makecell*[c]{Pooling} &Acc.\\
        \midrule
        Max &72.2 \\
        Ave &65.7 \\
        \textbf{Both} &\textbf{76.9} \\
	\bottomrule
	\end{tabular}
\end{adjustbox}
\end{minipage}
\label{supp_t1}
\end{table}

\paragraph{Point-PN.}
For the parametric variant, we decrease $C_I$ to 36 and neighbors number $k$ to 40 to achieve lightweight parameters and efficient inference. We adopt the bottleneck architecture with a ratio 0.5 for the two cascaded linear layers after the local geometry aggregation module. The initial parametric raw-point embedding consists of only one linear layer, and the final classifier contains three linear layers. For shape classification, we train Point-PN for 300 epochs with a batch size 32 on a single RTX 3090 GPU, and utilize the same data augmentation with 1,024 input points as previous works~\cite{pointmlp,qian2022pointnext}. On ModelNet40~\cite{modelnet40}, we adopt SGD optimizer with a weight decay 0.0002, and cosine scheduler with an initial learning rate 0.1. On ScanObjectNN~\cite{scanobjectnn}, we adopt AdamW optimizer~\cite{kingma2014adam} with a weight decay 0.05, and cosine scheduler with an initial learning rate 0.002. For part segmentation, we utilize the same decoder architecture and training settings as CurveNet~\cite{curvenet} for a fair comparison.

\begin{table}[t]
\centering
\caption{\textbf{Magnitude $\alpha$ in Trigonometric Functions.} We report the classification accuracy of Point-NN on ModelNet40~\cite{modelnet40}.}
\begin{adjustbox}{width=0.96\linewidth}
	\begin{tabular}{lcccccc}
	\toprule
	\makecell*[c]{Magnitude $\alpha$} &\ 1\ &10 &50 &\textbf{100} &200 &500\\
        \cmidrule(lr){1-1} \cmidrule(lr){2-7}
        \specialrule{0em}{1pt}{1pt}
		 Acc. (\%) &\ 52.7\ &65.3  & 74.2  &\textbf{76.9} &75.4 &74.8\ \\ 
		 \specialrule{0em}{1pt}{1pt}
	\bottomrule
	\end{tabular}
\end{adjustbox}
\label{supp_t2}
\end{table}\vspace{0.1cm}

\begin{table}[t]
\centering
\caption{\textbf{Wavelength $\beta$ in Trigonometric Functions.} We report the classification accuracy of Point-NN on ModelNet40~\cite{modelnet40}.}
\begin{adjustbox}{width=\linewidth}
	\begin{tabular}{lcccccc}
	\toprule
	\makecell*[c]{Wavelength $\beta$} &\ 10\ &100 &\textbf{500} &1000 &2000 &3000\ \\
        \cmidrule(lr){1-1} \cmidrule(lr){2-7}
        \specialrule{0em}{1pt}{1pt}
		 Acc. (\%) &\ 41.5\  &74.2 &\textbf{76.9} &74.5 &73.1 &72.9\ \\ 
		 \specialrule{0em}{1pt}{1pt}
	\bottomrule
	\end{tabular}
\end{adjustbox}
\label{supp_t3}
\vspace{0.1cm}
\end{table}

\section{Additional Ablation Study}
\label{s3}

\paragraph{Non-Parametric Encoder.}
In Table~\ref{supp_t1}, we further investigate other designs at every stage of Point-NN's non-parametric encoder. As shown, $k$-NN performs better than ball query~\cite{qi2017pointnet++} for grouping the neighbors of each center point since the ball query would fail to aggregate valid geometry in some sparse regions with only a few neighboring points.
Expanding the feature dimension by concatenating the center and neighboring points can improve the performance by +5.3\%. This is because each point obtains larger receptive fields as the network stage goes deeper and requires higher-dimensional vectors to encode more spatial semantics.  For the pooling operation after geometry extraction, we observe applying both max and average pooling achieves the highest accuracy, which can summarize the local patterns from two different aspects.

\begin{table}[t]
\centering
\begin{adjustbox}{width=\linewidth}
	\begin{tabular}{c c c cc c}
	\toprule
		\makecell*[c]{Raw\\Embed.} &\makecell*[c]{Linear\\Layers} &\makecell*[c]{Classifier} &\makecell*[c]{ModelNet40} &\makecell*[c]{ScanObjNN}
		&\makecell*[c]{Param.}\\
		 \cmidrule(lr){1-3} \cmidrule(lr){4-4} \cmidrule(lr){5-5} \cmidrule(lr){6-6}
	   N &-  &N & 76.9 & 61.5 &0.0 M\\
	     N &-  &P & 90.3 &68.5 &0.2 M\\
	     P &-  &P & 90.8 &73.7 &0.2 M\\
	     P &0+1  &P & 93.4 & 86.0 &0.5 M\\
          P &1+0  &P & 93.0 & 83.8 &0.5 M\\
          P &1+1  &P & 93.2 & 86.5 &0.7 M\\
          P &0+2  &P & 93.0 & 84.6 &0.7 M\\
          P &2+0  &P & 92.7 & 84.9 &0.7 M\\
	     P &2+1  &P & 92.7 & 83.6 &0.8 M\\
          P &2+2  &P & 92.9 & 82.7 &1.0 M\\
          \bf P &\bf 1+2  &\bf P & \bf 93.8 &\bf 87.1 &\bf 0.8 M\\
	\bottomrule
	\end{tabular}
\end{adjustbox}
\caption{\textbf{Detailed Step-by-step Construction of Point-PN} on ModelNet40~\cite{modelnet40} and ScanObjectNN~\cite{scanobjectnn}. We report the classification accuracy (\%) on the PB-T50-RS split of ScanObjectNN. `N' or `P' denotes the non-parametric modules or parametric linear layers. `Linear Layers' denotes the number of linear layers inserted into each network stage.}
\label{supp_t5}
\end{table}

\paragraph{Hyperparameters in Trigonometric Functions.}
In Table~\ref{supp_t2} and \ref{supp_t3}, we show the influence of two hyperparameters in trigonometric functions of Point-NN. We fix one of them to be the best-performing value ($\alpha$ as 100, $\beta$ as 500), and vary the other one for ablation.
The combination of the magnitude $\alpha$ and wavelength $\beta$ control the frequency of the channel-wise sinusoid, and thus determine the raw point encoding for different classification accuracy.

\paragraph{Point-Memory Bank with Different Sizes.}
As default, we construct the feature memory by the entire training-set point clouds. In Table~\ref{supp_t4}, we report how Point-NN performs when partial training samples are utilized for the point-memory bank. As shown, Point-NN can attain 60.1\% classification accuracy with only 10\% of the training data, and further achieves 70.1\% with 40\% data, which is comparable to the performance of 100\% ratio but consumes less GPU memory. This indicates Point-NN is not sensitive to the memory bank size and can perform favorably with partial training-set data.

\paragraph{Point-PN.}
In Table~\ref{supp_t5}, we present the detailed results of constructing Point-PN on ModelNet40~\cite{modelnet40} and ScanObjectNN~\cite{scanobjectnn}. As shown, the linear layers after the geometry aggregation module (`0+1', `0+2', `1+2') are more important than the previous ones (`1+0', `2+0', `2+1').
Also, inserting layers into both positions (`1+1') performs better than only into one position (`0+1', `1+0').

\paragraph{Memory Cost by Plug-and-play.}
In Table~\ref{supp_t44}, we present the GPU memory cost brought by the ensemble of Point-NN. As shown, utilizing Point-NN as a plug-and-play module brings no extra learnable parameters and causes marginal memory consumption, indicating its efficiency for training-free performance enhancement.

\begin{table}[t]
\centering
\caption{\textbf{Point-Memory Bank with Different Sizes.} We randomly sample different ratios of ModelNet40~\cite{modelnet40} to construct the point-memory bank and report the classification accuracy with GPU memory consumption.}
\begin{adjustbox}{width=\linewidth}
	\begin{tabular}{lccccccc}
	\toprule
	\makecell*[c]{Ratio (\%)} &1 &5 &10 &20 &40 &80 &100\\
        \cmidrule(lr){1-1} \cmidrule(lr){2-8}
        \specialrule{0em}{1pt}{1pt}
		 Acc. (\%)&37.9 &55.6  &60.1  &65.5 &70.1 &74.2 &\textbf{76.9}\\ 
		 \specialrule{0em}{1pt}{1pt}
         Mem. (G) &3.84 &3.87 &3.93  &4.05 &4.26 &4.82 &5.21\\ 
		 \specialrule{0em}{1pt}{1pt}
	\bottomrule
	\end{tabular}
\end{adjustbox}
\label{supp_t4}
\end{table}

\begin{table}[t]
\centering
\caption{\textbf{Memory Cost by Plug-and-play.} We respectively report the classification accuracy (\%) on ModelNet40~\cite{modelnet40}, learnable parameters, and GPU memory (G). All compared methods are conducted with batch size 32 on a single RTX 3090 GPU.}
\begin{adjustbox}{width=\linewidth}
	\begin{tabular}{lcccc}
	\toprule
	\makecell*[c]{Method} &+NN &Acc. (\%) &Param. &Mem. (G)\\
        \cmidrule(lr){1-1} \cmidrule(lr){2-5}
        \specialrule{0em}{1pt}{1pt}
		 Point-NN&- &76.9  &0.0 M  &5.21 \\ 
		 \specialrule{0em}{1pt}{1pt}
         Point-PN &- &93.8 &0.8 M  &6.63\\ 
         \midrule\specialrule{0em}{1pt}{3pt}
        PointNet++~\cite{qi2017pointnet++} &- &92.6 &1.7 M  &7.46\\
         &\checkmark &\textcolor{blue}{+0.5} &\textcolor{blue}{+0 M}  &\textcolor{blue}{+2.45} \\
         \specialrule{0em}{1pt}{3pt}
        PCT~\cite{guo2021pct} &- &93.2 &-  &5.59\\
         &\checkmark &\textcolor{blue}{+0.2} &\textcolor{blue}{+0 M}  &\textcolor{blue}{+3.25} \\
        \specialrule{0em}{1pt}{1pt}
        PointMLP~\cite{pointmlp} &- &94.1 &12.6 M  &19.07 \\
         &\checkmark &\textcolor{blue}{+0.3} &\textcolor{blue}{+0 M}  &\textcolor{blue}{+0.07} \\
	\bottomrule
	\end{tabular}
\end{adjustbox}
\label{supp_t44}
\end{table}

\begin{figure*}[t!]
  \centering
    \includegraphics[width=1\textwidth]{cvpr2023-author_kit-v1_1-1/latex/figs/supp_f1.pdf}
   \caption{\textbf{Why Do Trigonometric Functions Work?} For an input point cloud, we visualize its low-frequency and high-frequency geometries referring to~\cite{xu2021learning}, and compare with the feature responses of Point-NN, where darker colors indicate higher responses. As shown, Point-NN can focus on the high-frequency 3D structures with sharp variations of the point cloud.}
    \label{supp_f1}
\end{figure*}

\section{Discussion}
\label{s4}

\subsection{Why Do Trigonometric Functions Work?}

We leverage the trigonometric function to conduct non-parametric raw-point embedding and geometry extraction. It can reveal the 3D spatial patterns benefited from the following three properties.

\paragraph{Capturing High-frequency 3D Structures.}
As discussed in Tancik et al.~\cite{tancik2020fourier}, transforming low-dimensional input by sinusoidal mapping helps MLPs to learn the high-frequency content during training. Similarly to our non-parametric encoding, Point-NN utilizes trigonometric functions to capture the high-frequency spatial structures of 3D point clouds, and then recognize them from these distinctive characteristics by the point-memory bank. In Figure~\ref{supp_f2}, we visualize the low-frequency (Top) and high-frequency (Middle) geometry of the input point cloud, and compare them with the feature responses of Point-NN (Bottom). The high-frequency geometries denotes the spatial regions of edges, corners, and other fine-grained details, where the local 3D coordinates vary dramatically, while the low-frequency structure normally includes some flat and smooth object surfaces with gentle variations. As shown, aided by trigonometric functions, our Point-NN can concentrate well on these high-frequency 3D patterns.

\paragraph{Encoding Absolute and Relative Positions.}
Benefited from the nature of sinusoid, the trigonometric functions can not only represent the absolute position in the embedding space, but also implicitly encode the relative positional information between two 3D points. For two points, $p_i=(x_i, y_i, z_i)$ and $p_j=(x_j, y_j, z_j)$, we first obtain their $C$-dimensional embeddings referring to Equation (5$\sim$7) in the main paper, formulated as
\begin{align}
\operatorname{PosE}(p_i) &= \operatorname{Concat}(f^{x}_i,\ f^{y}_i,\ f^{z}_i),\\
\operatorname{PosE}(p_j) &= \operatorname{Concat}(f^{x}_j,\ f^{y}_j,\ f^{z}_j),
\end{align}
where $\operatorname{PosE}(\cdot)$ denotes the positional encoding by trigonometric functions, and  $f^x_{i/j},\ f^y_{i/j},\ f^z_{i/j} \in \mathbb{R}^{1\times \frac{C}{3}}$ denote the embeddings of three axes.
Then, their spatial relative relation can be revealed by the dot production between the two embeddings, formulated as
\begin{align}
f^{x}_i f^{x}_j^T + f^{y}_i f^{y}_j^T + f^{z}_i f^{z}_j^T = \operatorname{PosE}(p_i) \operatorname{PosE}(p_j)^T,\nonumber
\end{align}
Taking the x axis as an example,
\begin{align}
\sum_{m=0}^{\frac{C}{6}-1}\mathrm{cosine}(\alpha (x_i - x_j)/{\beta^{\frac{6m}{C}}}) = f^{x}_i f^{x}_j^T,
\end{align}
which indicates the relative x-axis distance of two points, in a similar way to the other two axes. Therefore, the trigonometric function is capable of encoding both absolute and relative 3D positional information for point cloud analysis.

\paragraph{Local Geometry Extraction.}

With the analysis above, we can reinterpret the Equation (9) of reweighing in the main paper from a mathematical perspective, which can be formulated as
\begin{align}
\label{weigh}
    f^w_{cj} &= \big(f_{cj} + \operatorname{PosE}(\Delta p_j)\big) \odot \operatorname{PosE}(\Delta p_j)\\
    &= f_{cj} \odot \operatorname{PosE}(\Delta p_j) + \operatorname{PosE}(\Delta p_j)\big \odot \operatorname{PosE}(\Delta p_j),\nonumber
\end{align}
where $\odot$ denotes element-wise multiplication, and $\Delta p_j, f_{cj}$ denote the relative coordinates and feature of neighbor point $j$ to the center point $c$. Now, there are two terms in Equation~\ref{weigh}, both of which reveal the relative geometry between point $i$ and $j$. The former implicitly captures the relative positional information from point $j$ to $i$, since the summation of its elements corresponds to the dot production between embeddings. The latter also indicates the embedding-space distance between the two points, which can be reformulated as the vector length of $\operatorname{PosE}(\Delta p_j)$. In this way, Point-NN can effectively extract local 3D patterns and aggregate them by pooling operations.

\begin{figure*}[t!]
  \centering
  \vspace{0.1cm}
    \includegraphics[width=0.94\textwidth]{cvpr2023-author_kit-v1_1-1/latex/figs/supp_f2.pdf}
   \caption{\textbf{Can Point-NN Improve Point-PN by Plug-and-play?} We visualize the feature responses for Point-NN, the trained PointNet++~\cite{qi2017pointnet++} and Point-PN, where darker colors indicate higher responses. As shown, Point-PN captures similar 3D patterns to Point-NN, which harms their complementarity.}
    \label{supp_f2}
    \vspace{0.1cm}
\end{figure*}

\subsection{Point-Memory Bank vs. $k$-NN?}
Based on the already extracted point cloud features, our point-memory bank and $k$-NN algorithm both leverage the inter-sample feature similarity for classification, and require no training or learnable parameters, but are different from the following two aspects.

\paragraph{Soft Integration vs. Hard Assignment.}
As illustrated in Section (2.3) of the main paper, our point-memory bank regards the similarities $S_{cos}$ between the test point cloud feature and the feature memory, $F_{mem}$, as weights, which are adopted for weighted summation of the one-hot label memory, $T_{mem}$. This can be viewed as a soft label integration. Instead, $k$-NN utilizes $S_{cos}$ to search the $k$ nearest neighbors from the training set, and directly outputs the category label with the maximum number of samples within the $k$ neighbors. Hence, $k$-NN conducts a hard label assignment, which is less adaptive than the soft integration. Additionally, our point-memory bank can be accomplished simply by two matrix multiplications, which is more efficient for the hardware.

\begin{table}[t]
\centering
\caption{\textbf{Point-Memory Bank vs. $k$-NN.} `Top-$k$ PoM' denotes the point-memory bank with top-$k$ similarities. We utilize our non-parametric encoder to extract features amd report the classification accuracy (\%) on ModelNet40~\cite{modelnet40}, where `All' denotes 9,840 training samples.}
\begin{adjustbox}{width=\linewidth}
	\begin{tabular}{cccccccc}
	\toprule
	\makecell*[c]{$k$} &1 &10 &100 &500 &1000 &5000 &All\\
        \cmidrule(lr){1-1} \cmidrule(lr){2-8}
        \specialrule{0em}{1pt}{1pt}
		 Top-$k$ PoM &71.6 &75.3  &76.1  &76.5 &76.7 &76.8 &\textbf{76.9}\\
		 \specialrule{0em}{1pt}{1pt}
         $k$-NN &71.8 &71.2 &62.0  &41.7 &34.5 &9.2 &-\\ 
		 \specialrule{0em}{1pt}{1pt}
	\bottomrule
	\end{tabular}
\end{adjustbox}
\label{supp_t7}
\end{table}

\paragraph{All Samples vs. $k$ Neighbors.}
Our point-memory bank softly integrates the entire label memory, which takes the semantics of all training samples into account. In contrast, $k$-NN only involves the nearest $k$ neighbors and discards the category knowledge of other training samples. 

\begin{table}[t]
\centering
\caption{\textbf{Can Point-NN Improve Point-PN by Plug-and-play?} We utilize Point-NN as an inference-time enhancement module to improve the trained Point-PN by interpolating their predictions. We report the accuracy (\%) on the PB-T50-RS split of ScanObjectNN~\cite{scanobjectnn}, ModelNet40~\cite{modelnet40}, and ShapeNetPart~\cite{shapenetpart}.}
\begin{adjustbox}{width=0.9\linewidth}
	\begin{tabular}{cccc}
	\toprule
	\makecell*[c]{Dataset} &ScanObjectNN &ModelNet40 &ShapeNetPart\\
        \cmidrule(lr){1-1} \cmidrule(lr){2-4}
        \specialrule{0em}{1pt}{1pt}
		 Point-PN &87.1 &93.8  &86.6 \\
		 \specialrule{0em}{1pt}{1pt}
         +NN &\textcolor{blue}{+0.1} &\textcolor{blue}{+0.2} &\textcolor{blue}{+0.0}\\ 
		 \specialrule{0em}{1pt}{1pt}
	\bottomrule
	\end{tabular}
\end{adjustbox}
\label{supp_t8}
\end{table}

\paragraph{Performance Comparison.}
In Table~\ref{supp_t7}, based on the point cloud features ectracted by our non-parametric encoder, we implement the top-$k$ version of point-memory bank for comparison with $k$-NN, which only aggregates the label memory of the training samples with top-$k$ similarities. 
As the neighbor number $k$ increases, $k$-NN's performance is severely harmed due to its hard label assignment, while our point-memory bank attains the highest accuracy by utilizing all 9,840 samples for classification, indicating their different characters.

\subsection{Can Point-NN Improve Point-PN by Plug-and-play?}
Point-NN can provide complementary geometric knowledge and serve as a plug-and-play module to boost the existing learnable 3D models after training. Although Point-PN is also a learnable 3D network, the enhanced performance brought by Point-NN is marginal as reported in Table~\ref{supp_t8}. By visualizing feature responses in Figure~\ref{supp_f2}, we observe that the complementarity between Point-NN and Point-PN is much weaker than that between Point-NN and PointNet++~\cite{qi2017pointnet++}. This is because the non-parametric framework of Point-PN is mostly inherited from Point-NN, and can also capture high-frequency 3D geometries via trigonometric functions. Therefore, the learnable Point-PN extracts similar 3D patterns to Point-NN, which harms its plug-and-play performance.

\section{Anonymous Code Release}

For reproducibility, we anonymously release our codes in \url{https://anonymous.4open.science/r/Non-Parametric_Networks_for_3D_Analysis-F246}.

{\small
\bibliographystyle{ieee_fullname}
\bibliography{egbib}
}